\documentclass{article}

\usepackage{colt11e}
\usepackage[round,comma]{natbib}
\usepackage{times}
\usepackage{amsfonts}
\usepackage{amsmath}
\usepackage[psamsfonts]{amssymb}
\usepackage{latexsym}
\usepackage{color}
\usepackage{graphics}
\usepackage{enumerate}
\usepackage{amstext}
\usepackage{url}
\usepackage{epsfig}

\newtheorem{proposition}{Proposition}

\newcommand{\mat}[1]{{\mathbf #1}}

\newcommand{\I}{\mat{I}}

\newenvironment{proof*}{\noindent{\bf Proof:}}{}
\newcommand{\ignore}[1]{}

\newcommand{\dd}{\mathrm{d}}

\newcommand{\EE}{\mathrm{E}}

\newcommand{\Real}{\mathbb{R}}

\newcommand{\fhat}{\hat{f}}
\newcommand{\gstar}{g^*}
\newcommand{\fstar}{f^*}
\newcommand{\Istar}{I_0}
\newcommand{\Jstar}{J_0}
\newcommand{\ftil}{\tilde{f}}
\newcommand{\Ihat}{\hat{I}}

\newcommand{\muone}{\mu_{1}}

\newcommand{\calH}{\mathcal{H}}

\newcommand{\calN}{\mathcal{N}}

\newcommand{\calX}{\mathcal{X}}

\newcommand{\Eqref}[1]{Eq.~{\eqref{#1}}}

\newcommand{\VCor}{V}
\newcommand{\Ctwo}{\tilde{V}}
\newcommand{\Diag}{\mathrm{Diag}}

\newcommand{\kmin}{\kappa}

\newcommand{\hnorm}[1]{\|_{\calH_{#1}}}

\newcommand{\lambdaone}{{\lambda_1^{(n)}}}
\newcommand{\lambdatwo}{{\lambda_2^{(n)}}}
\newcommand{\hSigma}{\hat{\Sigma}}
\newcommand{\tSigma}{\tilde{\Sigma}}

\newcommand{\pto}{\stackrel{p}{\to}}

\newcommand{\Id}{I_d}
\newcommand{\Jd}{J_d}

\newcommand{\LPi}{L_2(\Pi)}
\newcommand{\elltwo}{\|_{\ell_2}}
\newcommand{\ellone}{\|_{\ell_1}}

\newcommand{\elloneMKL}{block-$\ell_1$ MKL} 
\newcommand{\elltwoMKL}{block-$\ell_2$ MKL}
\newcommand{\ellpMKL}{block-$\ell_p$ MKL}
\newcommand{{\elasticMKL}}{elastic-net MKL}

\newcommand{{\ElasticMKL}}{Elastic-net MKL}

\newcommand{\Kconst}{K}

\newtheorem{Theorem}{Theorem}

\newtheorem{Definition}[Theorem]{Definition}
\newtheorem{Lemma}[Theorem]{Lemma}

\newtheorem{Assumption}{Assumption}

\title{Sharp Convergence Rate and Support Consistency of \\
Multiple Kernel Learning  with Sparse and Dense Regularization}
\author{Taiji Suzuki, Ryota Tomioka \\
Department of Mathematical Informatics, \\
The University of Tokyo,\\
7-3-1 Hongo, Bunkyo-ku, Tokyo\\
\texttt{\small t-suzuki@mist.i.u-tokyo.ac.jp}, \\ 
\texttt{\small tomioka@mist.i.u-tokyo.ac.jp}
\And
Masashi Sugiyama \\
Department of Computer Science, \\
Tokyo Institute of Technology,\\
2-12-1 O-okayama, Meguro-ku, Tokyo\\
\texttt{\small sugi@cs.titech.ac.jp}
}

\begin{document}

\maketitle

\begin{abstract}
We theoretically investigate
the convergence rate and support consistency (i.e., correctly identifying the subset of non-zero coefficients in the large sample limit) 
of multiple kernel learning (MKL).
We focus on MKL with block-$\ell_1$ regularization (inducing sparse kernel combination),
block-$\ell_2$ regularization (inducing uniform kernel combination),
and elastic-net regularization (including both block-$\ell_1$ and block-$\ell_2$ regularization).
For the case where the true kernel combination is sparse,
we show a sharper convergence rate of the block-$\ell_1$ and elastic-net MKL methods 
than the existing rate for block-$\ell_1$ MKL.
We further show that
elastic-net MKL requires a milder condition for being consistent than
block-$\ell_1$ MKL.
For the case where the optimal kernel combination is not exactly sparse,
we prove that elastic-net MKL can achieve a faster convergence rate 
than the block-$\ell_1$ and block-$\ell_2$ MKL methods by carefully controlling
the balance between the block-$\ell_1$and block-$\ell_2$ regularizers.
Thus, our theoretical results overall suggest the use of elastic-net
regularization in MKL.

\end{abstract}

\section{Introduction}




The choice of kernel functions is a key issue
for \emph{kernel methods} such as support vector machines to work well
\citep{book:Vapnik:1998}.
A traditional but very powerful approach to optimizing the kernel function is
the use of \emph{cross-validation} (CV) \citep{JRSS:Stone:1974}.
Although the CV-based kernel choice often leads to better generalization,
it is computationally expensive when the kernel contains multiple
tuning parameters.

To overcome this limitation, the framework of \emph{multiple kernel learning} (MKL)
has been introduced,
which tries to learn the optimal linear combination of prefixed base-kernels
by convex optimization \citep{JMLR:Lanckriet+etal:2004,JMLR:MicchelliPontil:2005,AS:Lin+Zhang:2005:COSSO,JMLR:Sonnenburg+etal:2006,JMLR:Rakotomamonjy+etal:2008,arXiv:Suzuki:2009}.
The seminal paper by \citet{ICML:Bach+etal:2004} showed that this MKL formulation
can be interpreted as block-$\ell_1$ regularization
(i.e., $\ell_1$ regularization across the kernels
and $\ell_2$ regularization within the same kernel).
We refer to this MKL formulation as `{\elloneMKL}'.
Based on this interpretation,
{\elloneMKL} was proved to be \emph{support consistent}
(i.e., correctly identifying the subset of non-zero coefficients with probability one
in the large sample limit)
when the true kernel combination is sparse
\citep{JMLR:BachConsistency:2008}.
Furthermore, the convergence rate of {\elloneMKL} has also been elucidated
in \cite{COLT:Koltchinskii:2008},
which can be regarded as an extension of the theoretical analysis for
ordinary (non-block) $\ell_1$ regularization
\citep{AS:Bickel+etal:2009,AS:TZhang:2009}.


However, in many practical applications,
the true kernel combination may not be exactly sparse.
In such a non-sparse situation, {\elloneMKL} was shown to perform rather poorly---just
the uniform combination of base kernels
obtained by block-$\ell_2$ regularization \citep{JMLR:MicchelliPontil:2005}
(which we call `{\elltwoMKL}')
often works better in practice \citep{ICMLtalk:Cortes:2009}. 
Furthermore, recent works showed
that some `intermediate' regularization between block-$\ell_1$ and block-$\ell_2$ regularization 
is more promising, e.g., block-$\ell_p$ regularization with $1\le p\le 2$
\citep{UAI:Cortes+etal:2009,NIPS:Marius+etal:2009},
and \emph{elastic-net} regularization \citep{JRSS:Zou+Hastie:2005}
which includes both block-$\ell_1$ and block-$\ell_2$ regularization
\citep{arXiv:SparsityTradeoff:2010} (we call this method `{\elasticMKL}').
Theoretically, the support consistency and the convergence rate for parametric elastic-nets 
have been elucidated in \cite{JRSS:YuanLin:2007} and \cite{AS:Zou+Zhang:2009}, respectively,
and that for non-parametric cases has been investigated
in \cite{AS:Meier+Geer+Buhlmann:2009}
focusing on the Sobolev space.

In this paper, we theoretically analyze
the support consistency and convergence rate of MKL,
and provide three new results.
\begin{itemize}
\item

For the case where the true kernel combination is sparse,
we show that {\elasticMKL} achieves a faster convergence rate than the one shown for
{\elloneMKL} \citep{COLT:Koltchinskii:2008}.
More specifically, 
we show that the $L_2$ convergence error is given by $\mathcal{O}_p(\min\{dn^{-\frac{2}{2+s}}+ d\log(M)/n,d^{\frac{1-s}{1+s}}n^{-\frac{1}{1+s}}+ d\log(M)/n\} )$,
where $d$ is the number of active components of the target function,
$s$ is the complexity of RKHSs, $M$ is the number of candidate kernels, and 
$n$ is the number of samples. 


\item 
For the case where the optimal kernel combination is not exactly sparse,
we prove that {\elasticMKL} achieves a faster convergence rate 
than the block-$\ell_1$ and block-$\ell_2$ MKL methods by carefully controlling
the balance between block-$\ell_1$ and block-$\ell_2$ regularization.
Our theoretical result well agrees with the experimental results reported in
\cite{arXiv:SparsityTradeoff:2010}.


\item 
For the case where the true kernel combination is sparse,
we prove that the necessary and sufficient conditions of
the support consistency
for {\elasticMKL} is milder than the conditions required for {\elloneMKL}
\citep{JMLR:BachConsistency:2008}.
\end{itemize}


Overall, our theoretical results suggest the use of elastic-net regularization in MKL.



\vspace{-2mm}

\section{Preliminaries}
\label{sec:Preliminary}
In this section, we formulate the {\elasticMKL} approach and 
summarize mathematical tools that are needed for the theoretical analysis.

\subsection{Formulation}
Suppose we are given $n$ samples $(x_i, y_i)_{i=1}^{n}$
where $x_i$ belongs
to an input space $\calX$ and $y_i \in \Real$. 
$(x_i,y_i)_{i=1}^n$ are independent and identically distributed from a probability measure $P$.
We denote the marginal distribution of $X$ by $\Pi$. 
We 
consider a MKL regression problem 
in which the unknown target function is represented as a form of 
$f(x)= \sum_{m=1}^M f_m(x)$,
where each $f_m$ belongs to different RKHSs
$\calH_m (m = 1,\dots,M)$ corresponding to
$M$ different base kernels $k_m$ over $\calX\times\calX$.


{\ElasticMKL} learns a decision function $\fhat$ as\footnote{
For simplicity, we focus on the squared-loss function here.
However, we note that it is straightforward to extend our convergence analysis and support consistency results
given in Sections~\ref{sec:ConvergenceRate} and \ref{sec:Consistency} to general loss functions
that are strongly convex and Lipschitz continuous,
by following the line of \cite{COLT:Koltchinskii:2008}. 
}
\begin{align}
\!\!\!\fhat = &\mathop{\arg \min}_{f_m \in \calH_m~(m=1,\dots,M)}
\frac{1}{n}\sum_{i=1}^n \left(y_i- \sum_{m=1}^M f_m(x_i) \right)^2
\!\!\! +
\lambdaone \sum_{m=1}^M \|f_m\hnorm{m} + \lambdatwo \sum_{m=1}^M \|f_m\hnorm{m}^2,
\!\!\!
\label{eq:primalElasticMKLnonpara}
\end{align}
where the first term is the squared-loss of function fitting
and, the second and the third terms are block-$\ell_1$ and block-$\ell_2$ regularizers,
respectively.
It can be seen from \eqref{eq:primalElasticMKLnonpara} that
{\elasticMKL} is reduced to {\elloneMKL} if $\lambdatwo=0$,
which tends to induce sparse kernel combination
\citep{JMLR:Lanckriet+etal:2004,ICML:Bach+etal:2004}.
On the other hand, it is reduced to {\elltwoMKL} if $\lambdaone=0$,
which results in uniform kernel combination \citep{JMLR:MicchelliPontil:2005}. 
It is worth noting that, {\elasticMKL} allows us to obtain various levels of
sparsity by controlling the ratio between $\lambdaone$ and $\lambdatwo$.

\subsection{Notations and Assumptions}

Here, we prepare technical tools needed in the following sections.  

Due to Mercer's theorem, 
there are an orthonormal system $\{\phi_{k,m}\}_{k,m}$ in $L_2(\Pi)$
and the spectrum $\{\mu_{k,m}\}_{k,m}$
such that $k_m$ has the following spectral representation: 
\begin{equation}
k_m(x,x') = \sum_{k=1}^{\infty} \mu_{k,m} \phi_{k,m}(x) \phi_{k,m}(x'). 
\label{eq:spectralRepre}
\end{equation}
By this spectral representation, the inner-product of RKHS can be expressed as 
$
\langle f_m ,g_m \rangle_{\calH_m} = \sum_{k=1}^{\infty} \mu_{k,m}^{-1} \langle f_m, \phi_{k,m} \rangle_{\LPi} \langle \phi_{k,m}, g_m \rangle_{\LPi}.
$

Let $\calH = \calH_1 \oplus \dots \oplus \calH_M$. For $f =(f_1,\dots,f_M) \in \calH$ and a subset of indices $I \subseteq \{1,\dots,M\}$, we 
denote by $f_I$ the restriction of $f$ to an index set $I$, i.e.,
$f_I = (f_m)_{m \in I}$. 

We denote by $\Istar$ the indices of truly active kernels, i.e.,
$$
\Istar=\{m \mid \|\fstar_m\hnorm{m}>0\},
$$
and define the complement of $\Istar$ as $\Jstar = {\Istar}^c$.

Throughout the paper, we assume the following technical conditions 
(see also \cite{JMLR:BachConsistency:2008}). 
\begin{Assumption}{\bf(Basic Assumptions)}\ 
\begin{enumerate}
\item[$\mathrm{(A1)}$]
There exists $\fstar = (\fstar_1,\dots,\fstar_M) \in \calH$
such that $\EE[Y|X] = \sum_{m=1}^M \fstar_m(X)$,
and the noise $\epsilon := Y - \fstar(X)$ has a strictly positive variance;
there exists $\sigma>0$ such that $\EE[\epsilon^2 | X] > \sigma^2 $ for all $X \in \calX$.
We also assume that $\epsilon$ is bounded as $|\epsilon| \leq L$.
\item[$\mathrm{(A2)}$]
For each $m=1,\dots,M$, $\calH_m$ is separable and $\sup_{X\in \calX} |k_m(X,X)| < 1$.
\item[$\mathrm{(A3)}$]
There exists $\gstar_m \in \calH_m$ such that 
\begin{equation}
\fstar_m(x) = \int_{\calX}k_m^{(1/2)}(x,x')\gstar_m(x')\dd \Pi(x')
 \qquad(\forall m = 1,\dots,M), \label{eq:fstarSigmacond}
\end{equation} 
where $k_m^{(1/2)}(x,x')= \sum_{k=1}^{\infty} \mu_{k,m}^{1/2}
  \phi_{k,m}(x) \phi_{k,m}(x')$ is the operator square-root of $k_m$.
\end{enumerate}
\end{Assumption}
The first assumption in (A1) ensures the model $\calH$ is correctly specified, 
and the technical assumption $|\epsilon| < L$ allows $\epsilon f$ to be Lipschitz continuous with respect to $f$.
  
It is known that the assumption (A2) gives the following relation: 
\begin{align*}
\|f_m\|_{\infty} \!\! \leq \! \sup_{x}\langle k_m(x,\cdot), f_m \rangle_{\calH_m}
\!\! \leq \! \sup_{x} \| k_m(x,\cdot)\hnorm{m}\! \|f_m\hnorm{m} 
\!\! \leq \!  
\sup_{x}  \sqrt{k_m(x,x)} \|f_m\hnorm{m} 
\! \leq \! \|f_m\hnorm{m}.
\end{align*}

\begin{table}[t]
\centering
\caption{Summary of the constants we use in this article.}
\label{tab:constants}
\begin{tabular}{|c|l|}
\hline
$M$ & The number of candidate kernels.  \\ \hline
$d$ & The number of active kernels of the truth; i.e., $d=|I_0|$. \\ \hline
$R$ & The upper bound of $\sum_{m=1}^M (\|\fstar_m \hnorm{m} +
     \|\fstar_m \hnorm{m}^2)$; see (A4). \\ \hline
$s$ & The spectral decay coefficient; see (A5). \\ \hline 
$\beta$ & The approximate sparsity coefficient; see (A7). \\ \hline
$b$ & The parameter that tunes the correlation between kernels; see (A8). \\ \hline
\end{tabular}
\end{table}

The assumption (A3) was used in \cite{FCM:Caponetto+Vito:2007} and also in \cite{JMLR:BachConsistency:2008}. 
It ensures the consistency of the least-squares estimates in terms of the RKHS norm.  
Using the spectral representation \eqref{eq:spectralRepre},
the condition $\gstar_m \in \calH_m$ is expressed as    
\begin{equation}
\|\gstar_m\hnorm{m}^2 =\sum_{k=1}^{\infty} \mu_{k,m}^{-2} \langle \fstar_m,\phi_{k,m} \rangle_{\LPi}^2 < \infty.
\end{equation}
This condition was also assumed in \cite{COLT:Koltchinskii:2008}. 
Proposition 9 of \cite{JMLR:BachConsistency:2008} gave
a sufficient condition to fulfill \eqref{eq:fstarSigmacond}
for translation invariant kernels $k_m(x,x') = h_m(x-x')$.

Constants we use later are summarized in Table~\ref{tab:constants}.

\section{Convergence Rate of {\ElasticMKL}}
\label{sec:ConvergenceRate}
In this section, we derive the convergence rate of {\elasticMKL}
in two situations:
\begin{enumerate}
  \item[(i)] A sparse situation where the truth $\fstar$ is sparse (Section~\ref{sec:Sparse}).
  \item[(ii)] A near sparse situation where the truth is not exactly sparse,
    but $\|f_m\hnorm{m}$ decays polynomially as $m$ increases (Section~\ref{sec:nearSparse}).
\end{enumerate}
For (i), we show that {\elasticMKL} (and {\elloneMKL}) achieves
a faster convergence rate than the rate shown for {\elloneMKL}~\citep{COLT:Koltchinskii:2008}. 
Furthermore, for (ii), 
we show that {\elasticMKL} can outperform {\elloneMKL} and {\elltwoMKL}
depending on the sparsity of the truth and the condition of the problem.
Throughout this section, we assume the following conditions.
\begin{Assumption}{\bf (Boundedness Assumption)}
There exists constants $C_1$ and $R$ such that 
\begin{flalign*}
\text{\rm(A4)} && \max_{m \in \Istar} \frac{\|\gstar_m \hnorm{m} }{\|\fstar_m \hnorm{m}} \leq C_1,~~~\sum\limits_{m=1}^M(\|\fstar_m\hnorm{m} + \|\fstar_m\hnorm{m}^2) \leq R. &&
\end{flalign*}
\end{Assumption}

\begin{Assumption}{\bf (Spectral Assumption)}
There exist $0 < s < 1$ and $C_2$ such that 
\begin{flalign*}
\text{\rm(A5)} &&
\mu_{k,m} \leq C_2 k^{-\frac{1}{s}},~~~(1\leq \forall k, 1\leq \forall m \leq M),&&
\end{flalign*}
where $\{\mu_{k,m}\}_{k}$ is the spectrum of the kernel $k_m$ (see Eq.\eqref{eq:spectralRepre}).
\end{Assumption}
The first assumption in (A4) appeared in Theorem 2 of \cite{COLT:Koltchinskii:2008}. 
The second assumption in (A4) bounds the amplitude of $\fstar$.
It was shown that the spectral assumption (A5) is equivalent to 
the classical covering number assumption~\citep{COLT:Steinwart+etal:2009}.
Recall that 
the $\epsilon$-covering number $\calN(\epsilon,\mathcal{B}_{\calH_m},\LPi)$ with respect to $\LPi$
is the minimal number of balls with radius $\epsilon$ needed to cover the unit ball $\mathcal{B}_{\calH_m}$ in $\calH_m$ \citep{Book:VanDerVaart:WeakConvergence}.
If the spectral assumption (A5) holds, there exists a constant $c$ that
depends only on $s$ such that 
\begin{align}
\label{eq:coveringcondition}
\calN(\varepsilon,\mathcal{B}_{\calH_m},\LPi) \leq c \varepsilon^{-2 s},
\end{align}
and the converse is also true (see Theorem 15 of \cite{COLT:Steinwart+etal:2009} and \cite{Book:Steinwart:2008} for details).
Therefore, 
if $s$ is large, 
at least one RKHS is ``complex'',
and if $s$ is small, the RKHSs are regarded as ``simple''.

For a given set of indices $I \subseteq \{1,\dots, M \}$, 
let $\kmin(I)$ be defined as follows:
\begin{align*}
\kmin(I) &:= \sup\left\{\kappa \geq 0 \mid \kappa \leq 
\frac{\|\sum_{m\in I}f_m\|_{\LPi}^2}{\sum_{m\in I}\|f_m\|_{\LPi}^2} ,~\forall f_m \in \calH_m~(m\in I)\right\}. 
\end{align*}
$\kappa(I)$ represents the correlation of RKHSs inside the
 indices $I$.  Similarly, we define the correlations of RKHSs between $I$ and $I^c$ as follows:
\begin{align*}
\rho(I) &:= \sup \left\{\frac{\langle f_I, g_{I^c} \rangle_{\LPi} }{\|f_I\|_{\LPi}\|g_{I^c} \|_{\LPi}} 
\mid f_I \in \calH_I, g_{I^c} \in \calH_{I^c}, f_I \neq 0, g_{I^c} \neq 0 \right\}. 
\end{align*}
In Subsections \ref{sec:Sparse} and \ref{sec:nearSparse}, 
we will assume that the kernels have {\it no perfect canonical dependence},  
implying that the kernels are not similar to each other (see (A6) 
and (A8) below). 

Throughout this paper, we assume 
$
\frac{\log(Mn)}{n} \leq 1$
and 
$\log(M)$ is slower than any polynomial order against the number of samples $n$: 
$\log(M) = o(n^{\epsilon})$ for all $\epsilon>0$.
With some abuse,
we use $C$ to denote constants that are independent of $d$ and $n$;
its value may be different.

\subsection{Sparse Situation}
\label{sec:Sparse}

Here we derive the convergence rate of the estimator $\fhat$ when the truth $\fstar$ is sparse.
Let $d = |\Istar |$
and suppose that the number of kernels $M$ and
the number of active kernels $d$ are increasing
with respect to the number of samples $n$. 
We further assume the following condition in this subsection.
\begin{Assumption}{\bf (Incoherence Assumption)}
There exists a constant $C_3>0$ such that 
\begin{flalign}
\text{\rm(A6)} && 0 <C_3^{-1} < \kmin(\Istar)(1-\rho^2(\Istar)). &&
\end{flalign}
\end{Assumption}
This condition is known as the {\it incoherence condition} \citep{COLT:Koltchinskii:2008,AS:Meier+Geer+Buhlmann:2009},
i.e., kernels are not too dependent on each other and the problem is well conditioned.
Then we have the following convergence rate. 
\begin{Theorem}
\label{th:SparseRate}
Under assumptions (A1-A6),  
there exist constants $C$, $F$ and $K$ depending only on 
$\kmin(\Istar)$, 
$\rho(\Istar)$, $s$, $C_1$, $C_2$, $L$, and $R$ such that 
the $\LPi$-norm of the residual $\fhat - \fstar$ can be bounded as follows:
when 
$d^{3+s}n^{-1} \leq 1$, for 
$\lambdaone = \lambdatwo = \max\{\Kconst n^{-\frac{1}{2+s}} + \tilde{K}_2 \sqrt{\frac{t}{n}},  F \sqrt{\frac{\log(M n) }{n}} \}$, 
\begin{align}
& \|\fhat - \fstar \|_{\LPi}^2 
\leq 
C \Big( d n^{-\frac{2}{2+s}} +  \frac{d t}{n} \Big), 
\label{eq:boundForL2Sparse}
\end{align}
and, when $d^{3+s}n^{-1} > 1$, 
for $\lambdaone  =  \max\{\Kconst (1+\sqrt{t})n^{-\frac{1}{2}},  F \sqrt{\frac{\log(M n) }{n}} \}$ and $\lambdatwo \leq \lambdaone$,
\begin{align}
& \|\fhat - \fstar \|_{\LPi}^2 
\leq 
C \Big( d^{\frac{1-s}{1+s}} n^{-\frac{1}{1+s}} + \frac{d (\log(M n) + t)}{n} \Big), 
\label{eq:boundForL2Sparse2}
\end{align}
where each inequality holds with probability at least $1 - e^{-t} - n^{-1}$ for all $t \geq \log \log (R\sqrt{n}) + \log M$.

\end{Theorem}
The above theorem indicates that 
the learning rate depends on the complexity of RKHSs (the simpler, the faster) and the number of
\emph{active} kernels rather than the number of kernels $M$
(the influence of $M$ is at most $\frac{d \log(M)}{n}$).
It is worth noting that the convergence rate in \eqref{eq:boundForL2Sparse} and \eqref{eq:boundForL2Sparse2} 
is faster than or equal to the rate of {\elloneMKL} shown by \cite{COLT:Koltchinskii:2008} which established the learning rate 
$O_p\left(d^{\frac{1-s}{1+s}} n^{- \frac{1}{1+s}} + \frac{d\log(M)}{n}\right)$ under the same conditions as ours 
\footnote{In our second bound \eqref{eq:boundForL2Sparse2}, 
there is the additional $\frac{d \log(n)}{n}$ term. 
However this can be eliminated by replacing the probability $1 - e^{-t} - n^{-1}$ with $1 - e^{-t} - M^{-A}$ as in \cite{COLT:Koltchinskii:2008}.
Moreover, if $\sqrt{n}\log(n)^{-\frac{1+s}{2s}} \geq d$, then the term $\frac{d \log(n)}{n}$ is dominated by the first term $ d^{\frac{1-s}{1+s}} n^{-\frac{1}{1+s}}$.}.

\subsection{Near-Sparse Situation}
\label{sec:nearSparse}

In this subsection, 
we analyze the convergence rate under a situation where 
$\fstar$ is not sparse but {\it near sparse}.
We have shown a faster learning rate than existing bounds in the previous subsection. 
However, the assumptions we used might be too restrictive to
capture the situation where MKL is used in practice. In fact, it was pointed
out in \cite{JRSS:Zou+Hastie:2005}
in the context of (non-block) $\ell_1$ regularization that $\ell_1$
regularization could fail in the following situations:
\begin{itemize}

\item When the truth $\fstar$ is not sparse, the $\ell_1$ regularization shrinks many small but non-zero components to zero.


\item When there exist strong correlations between different kernels,
  the solution of {\elloneMKL} becomes unstable.


\item When the number of kernels $M$ is not large,
  there is no need to impose the estimator to be sparse.
\end{itemize}


In order to analyze these situations in the MKL setting, we introduce three
parameters $\beta$, $b$, and $\tau$:
$\beta$ controls the level of sparsity (see (A7)), $b$ controls the
correlation between candidate kernels (see (A8)), and $\tau$ controls the
growth of the number of kernels against the number of samples (see (A9)).

We show that naturally {\elltwoMKL} is preferable when there are
only few candidate kernels or the truth is dense. Importantly, if the
candidate kernels are correlated, the convergence of {\elloneMKL}
can be slow even when the truth is sparse. Our analysis shows
that {\elasticMKL} is most valuable in such an intermediate situation.




By permuting indices, we can assume without loss of generality that 
$\|\fstar_m \hnorm{m}$ is decreasing with respect to $m$, 
i.e., $\|\fstar_1 \hnorm{1} \geq \|\fstar_2 \hnorm{2} \geq \|\fstar_3 \hnorm{3} \geq \cdots$.
We further assume the following conditions in this subsection.
\begin{Assumption}{\bf (Approximate Sparsity)}
\label{ass:nonsparse}
The truth is approximately
 sparse, i.e., 
$\|\fstar_m \hnorm{m} > 0 $ for all $m$ and thus $\Istar =
 \{1,\dots,M\}$. However,
$\|\fstar_m\hnorm{m}$ decays polynomially with respect to
$m$ 
as follows:
\begin{flalign*}
{\rm (A7)} && \|\fstar_m \hnorm{m} \leq C_3 m^{-\beta}. &&
\end{flalign*}
We call $\beta~(>1)$ the {\it approximate sparsity coefficient}.
\end{Assumption}

\begin{Assumption}{\bf (Generalized Incoherence)}
There exist $b>0$ and $C_4$ such that for all $I \subseteq \{1,\dots,M\}$, 
\begin{flalign*}
{\rm (A8)} && (1-\rho^2(I))\kmin(I) \geq C_4 |I|^{-b}. &&
\end{flalign*}
\end{Assumption}
\begin{Assumption}{\bf (Kernel-Set Growth)}
The number of kernels $M$ is increasing polynomially with respect to the number of samples $n$, i.e., 
$\exists \tau > 0$ such that
\begin{flalign*}
{\rm (A9)}&&  
M = \lceil n^{\tau} \rceil.
&&
\end{flalign*}
\end{Assumption}


For notational convenience, let 
$\tau_1=\frac{1}{(2\beta+b)(2+s)-1-s}$, 
$\tau_2=\frac{(s-1)(2\beta-1) + bs}{(2\beta+b)(2+s)-1-s}$,  
$\tau_3=\frac{s\{2(b+\beta)-1\}}{2(2+s)(b+\beta)-s}$, 
$\tau_4=\frac{s}{2+s}$,
$\tau_5=\frac{b+1}{(\beta+b)\{b(2+s)+2\}}$ and 
$\tau_6=\frac{1}{(1-s)(1+b)}$. 
In addition, we denote by $\Kconst$ some sufficiently large constant.


\begin{Theorem}
\label{th:NearSparse}
Suppose assumptions (A1-A5) and (A7-A9),
$2\beta(1-s) < s(b-1) $, and 
$\tau_1 <  \tau < \tau_4$ are satisfied. 
Then the estimator of {\elasticMKL} possesses the following convergence rate
each of which holds with probability at least $1 - e^{-t} - n^{-1}$ for all $t \geq \log\log (R\sqrt{n}) + \log M$: \\
\noindent 1. When $\tau_1 <  \tau < \tau_2$, 
\begin{align}
\label{eq:elastconvone}
 \|\fhat - \fstar \|_{\LPi}^2 
\leq 
&C \Big\{  n^{- \gamma_1 } +  
(n^{-\frac{(2\beta+b)(2+s)-3-s+2\beta}{2\{(2\beta+b)(2+s)-1-s\}}} + \lambdatwo^2)(\sqrt{t}+t)
 \Big\}, \notag \\
&~~\text{where}~~\gamma_1 = \frac{ 4\beta +b- 2}{(2+s)(2\beta+b)-1-s}, &&&&
\end{align}
with $\lambdaone =  \max\{\Kconst n^{-\frac{3\beta + b -1}{(2\beta + b)(2+s)-1-s}} + \tilde{K}_2\sqrt{\frac{t}{n}},  F \sqrt{\frac{\log(M n) }{n}} \} $ and 
$\lambdatwo = \Kconst n^{-\frac{2\beta + b -1}{(2\beta + b)(2+s)-1-s}}$.\\
\noindent 2. When $\tau_2 \leq \tau < \tau_3$,  
\begin{align}
\label{eq:elastconvtwo}
 \|\fhat - \fstar \|_{\LPi}^2 
\leq 
&C \Big\{ n^{\tau\frac{(2+s)b+2}{2\{(2+s)(b+\beta)-s\}}-\gamma_2} +  (n^{\frac{\tau(2+s)(1-\beta) -(4\beta+2b+sb-2)}{2\{(\beta+b)(2+s)-s\}}}+\lambdatwo^2)(\sqrt{t} + t) \Big\}, \notag \\
&~~\text{where}~~\gamma_2 = \frac{4\beta + b(2+s) -2}{2\{(2+s)(b+\beta)-s\}}, 
\end{align}
with  $\lambdaone = \max\{\Kconst \sqrt{\frac{M}{n}} + \tilde{K}_2\sqrt{\frac{t}{n}},  F \sqrt{\frac{\log(M n) }{n}} \} $ and 
$\lambdatwo = \Kconst  n^{\frac{\tau - \{2(b+\beta) -1 \}}{2\{(2+s)(b+\beta)-s\}}}$.

\noindent 3. When $\tau_3 \leq \tau < \tau_4$,  
\begin{align}
\label{eq:elastconvthree}
 \|\fhat - \fstar \|_{\LPi}^2 
\leq &
C \Big( n^{\tau\gamma_3 - \gamma_3} +  (n^{\frac{\tau(\beta-1) + 1-2\beta-b}{2(b+\beta)}} + \lambdatwo^2)(\sqrt{t} + t) \Big), \notag \\
&~~\text{where}~~\gamma_3 = \frac{b+2\beta-1}{2 (b+\beta)}, 
\end{align}
with  $\lambdaone = \max\{\Kconst \sqrt{\frac{M}{n}} + \tilde{K}_2\sqrt{\frac{t}{n}},  F \sqrt{\frac{\log(M n) }{n}} \}$ and 
$\lambdatwo = \Kconst (M/n)^{\frac{2(b+\beta)-1}{4(b+\beta)}}$.

\end{Theorem}

\begin{Theorem}
\label{th:NearSparseL1L2}
Under assumptions (A1-A5) and (A7-A9),
if $\tau_5 <\tau$, 
the estimator $\fhat_{\ell_1}$ of {\elloneMKL} has the following convergence rate
with probability at least $1 - e^{-t} - n^{-1}$ for all $t \geq \log\log (R\sqrt{n}) + \log M$:
\begin{flalign}
(\text{\elloneMKL})&&
\|\fhat_{\ell_1} - \fstar \|_{\LPi}^2 \leq C\left(  n^{-\gamma_4} 
 +  n^{-\frac{4\beta+2b-2+s(b+\beta)}{2(2+s)(b+\beta)}}(\sqrt{t} + t)  \right), \notag  \\
&&~~\text{where}~~\gamma_4 =  \frac{2\beta + b-1}{(\beta+b)(2+s)}, &&
\end{flalign}
with $\lambdaone = \max\{ \Kconst n^{-\frac{1}{2+s}} + \tilde{K}_2\sqrt{\frac{t}{n}},  F \sqrt{\frac{\log(M n) }{n}} \}$ and $\lambdatwo = 0$. 
Moreover, if $\tau < \tau_6 $, the estimator $\fhat_{\ell_{2}}$ of {\elltwoMKL} has the following convergence rate
with probability at least $1 - e^{-t} - n^{-1}$ for all $t \geq \log\log (R\sqrt{n}) + \log M$:
\begin{flalign}
(\text{\elltwoMKL})
&&
\|\fhat_{\ell_2} - \fstar \|_{\LPi}^2 \leq C\left( n^{\tau(b+\frac{2}{2+s})-\gamma_5}   
 + \left(\lambdatwo^2 + \frac{M^{1+b}}{n}\right)t \right),&& \notag \\
&&~~\text{where}~~\gamma_5 =  \frac{2}{2+s}, 
&&
\end{flalign}
with $\lambdatwo = \max\{\Kconst (\frac{M}{n})^{\frac{1}{2+s}},  F \sqrt{\frac{\log(M n) }{n}} \}$ and $\lambdaone = 0$.
\end{Theorem}


In all convergence rates presented in Theorems \ref{th:NearSparse} and \ref{th:NearSparseL1L2},
the leading terms are the terms that do not contain $t$.
The convergence order of the terms containing $t$ are faster than the leading terms, thus negligible.

\begin{figure}[t]
 \centering
 \includegraphics[width=.5\textwidth]{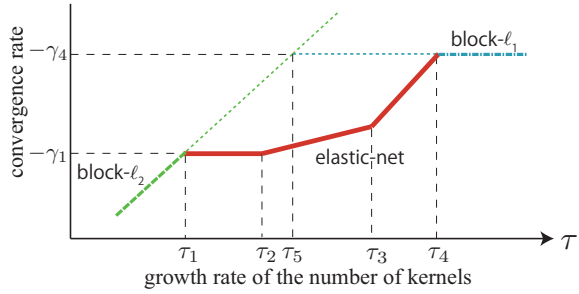}
 \caption{
  Relation between the convergence rate and the number of kernels.
  If the truth is intermediately sparse (the growth rate $\tau$ of the number of kernels is between $\tau_1$ and $\tau_5$), 
then {\elasticMKL} performs best. 
At the edge of the interval, the convergence rate of {\elasticMKL} coincides with that of {\elloneMKL} or {\elltwoMKL}.}
 \label{fig:graph}
\end{figure}

By simple calculation, we can confirm that 
{\elasticMKL} always converges faster than {\elloneMKL} and {\elltwoMKL} 
if $\beta$ and $M$ satisfy the condition of Theorem  \ref{th:NearSparse}. 
The convergence rate of {\elasticMKL} becomes identical with {\elltwoMKL} and {\elloneMKL} 
at the two extreme points of the interval $\tau = {\tau_1}$ and ${\tau_4}$, respectively.
Outside the region, {\elloneMKL} or {\elltwoMKL} has a faster convergence rate
than \elasticMKL.
Moreover, at $\tau = \tau_2$, the convergence rates \eqref{eq:elastconvone}
and \eqref{eq:elastconvtwo} of {\elasticMKL} are identical, and at $\tau = \tau_3$,
the convergence rates \eqref{eq:elastconvtwo} and \eqref{eq:elastconvthree} are identical.
The relation between the most preferred method and the growth rate $\tau$ of the number of kernels is illustrated in Figure~\ref{fig:graph}.


The condition $\tau_1 < \tau < \tau_4$ 
in Theorem \ref{th:NearSparse} indicates that 
when the number of kernels is not too small or too large,
an `intermediate' effect of {\elasticMKL} becomes advantageous.
Roughly speaking, if $M$ is large,
sparsity is needed to ensure the convergence
and thus {\elloneMKL} performs the best.
On the other hand, if $M$ is small, 
there is no need to make the solution sparse
and thus {\elltwoMKL} becomes the best.
For an intermediate $M$, {\elasticMKL} becomes the best.

The condition $2\beta(1-s) < s(b-1)$ in Theorem \ref{th:NearSparse} ensures 
the existence of $M$ that satisfies the condition in the theorem, i.e.,
$\tau_1 < \tau_2 < \tau_3 < \tau_4$.
It can be seen that as $b$ becomes large (the condition of the problem becomes worse), 
the range of $\beta$ and $M$ 
in which {\elasticMKL} performs better than
{\elloneMKL} and {\elltwoMKL} becomes large.
This indicates that the worse the condition of the problem becomes,
the more important to control the balance of $\lambdaone$ and $\lambdatwo$ appropriately.

\section{Support Consistency of {\ElasticMKL}}
\label{sec:Consistency}

In this section, we derive necessary and sufficient conditions 
for the statistical support consistency of the estimated sparsity pattern, i.e., 
the probability of $\{m\mid \|\fhat_m \hnorm{m} \neq 0\} = I_0$ goes to 1
as the number of samples $n$ tends to infinity. 
Due to the additional squared regularization term, the necessary condition for the support consistency of {\elasticMKL} 
is shown to be weaker than that for {\elloneMKL} \citep{JMLR:BachConsistency:2008}.  
In this section, we assume $M$ and $d=|I_0|$ are fixed against the number of samples $n$.

Let $\calH_I$ be the restriction of $\calH_1 \oplus \dots \oplus
\calH_M$ to the index set $I$.
Since $\EE_{X}[k_m(X,X)] < \infty$ for all $m$ (from assumption (A2)),  we
define the (non-centered)  
{\it cross covariance operator} $\Sigma_{I,J}:\calH_{I} \to \calH_{J}$ 
 as a bounded linear operator such that\footnote{
If one fits a function with a constant offset ($f(x) +b$ instead of $f(x)$) as in \cite{JMLR:BachConsistency:2008},
then the centered version of cross covariance operator
is required instead of the non-centered version, i.e.,
$\langle f_m, \Sigma_{m, m'} g_{m'} \rangle_{\calH_m} = \EE_X[(f_m(X) - \EE_X[f_m])(g_{m'}(X) - \EE_X[g_{m'}])]$.
However, this difference is not essential because, without loss of generality,
one can consider a situation where $\EE_Y[Y] = 0$ and $\EE_X[f_m(X)]= 0$ for all $f_m \in \calH_M$
by centering all the functions.
}
\begin{align}
\langle f_I, \Sigma_{I, J} g_J \rangle_{\calH_I} = \sum_{m\in
 I}\sum_{m'\in J}\langle f_m, \Sigma_{m, m'} g_{m'} \rangle_{\calH_m}
=\sum_{m\in
 I}\sum_{m'\in J}
\EE_X[f_m(X)g_{m'}(X)], 
\label{eq:CCVop}
\end{align}
for all $f_I=(f_m)_{m\in I} \in \calH_I$ and $g_J=(g_{m'})_{m'\in J} \in \calH_J$.
See \cite{TAMS:Baker:1973} for the details of the cross covariance
operator $(f,g) \mapsto \mathrm{cov}(f(X)g(X))$. 

 Moreover, we define the bounded (non-centered) {\it cross-correlation operators}\footnote{
Actually, such a bounded operator always exists \citep{TAMS:Baker:1973}.
}
 $\VCor_{l,m}$ by 
$\Sigma_{l,l}^{1/2}\VCor_{l,m}\Sigma_{m,m}^{1/2} = \Sigma_{l,m}$.
The joint cross-correlation operator $\VCor_{I,J}:\calH_J\rightarrow \calH_I$ is defined analogously
to  $\Sigma_{I,J}$.

In this section, we assume in addition to the basic assumptions (A1-A3) that
\begin{enumerate}
 \item[$\mathrm{(A10)}$] All $\VCor_{l,m}$ are compact and the joint correlation operator $\VCor$ is invertible.
\end{enumerate}

Let $\Ihat$ be the indices of {\it active kernels}
for the estimated $\fhat \in \calH$ by {\elasticMKL}: $\Ihat := \{m \mid \|\fhat_m\hnorm{m} > 0\}$.
Let
$
D :=  \Diag(\|\fstar_m\hnorm{m}^{-1}) = \Diag((\|\fstar_m\hnorm{m}^{-1})_{m \in \Istar})
$,
where $\Diag$ is the $|\Istar| \times |\Istar|$ block-diagonal operator with operators $\|\fstar_m\hnorm{m}^{-1} \I_{\calH_m}$ on diagonal blocks for $m\in \Istar$.
In this section, we assume that the true sparsity pattern $\Istar$ and the number of kernels $M$ are fixed independently of the number of samples $n$.

The norm of $f\in \calH$ is defined by $\|f\hnorm{} := \sqrt{\sum_{m=1}^M \|f_m\hnorm{m}^2} $ and 
similarly that of $f_I \in \calH_I$ is defined by $\|f_I\hnorm{I} := \sqrt{\sum_{m\in I} \|f_m\hnorm{m}^2}$ .
The following theorem gives a sufficient condition for the support consistency of sparsity patterns.

\begin{Theorem}
\label{th:SufconsistencyElastMKL}
Suppose $\lambdatwo > 0$, $\lambdaone \to 0,$ $\lambdatwo \to 0$, $\lambdaone \sqrt{n} \to \infty$, 
and 
\begin{align}
&
\textstyle 
\limsup_n \left\|\Sigma_{m, \Istar} (\Sigma_{\Istar,\Istar} + \lambdatwo)^{-1}\left(D + 2 \frac{\lambdatwo}{\lambdaone}\right)f^*_{\Istar} \right\hnorm{m}  
< 1,~~~~~(\forall m \in J =  \Istar^c).
\label{eq:IRcondsuff}
\end{align}
Then\footnote{For random variables $x_n$ and $y$, 
  $x_n \stackrel{p}{\rightarrow} y$ means the convergence in probability, 
i.e., the probability $|x_n-y| > \epsilon$ goes to 0 for all $\epsilon$
as the number of samples $n$ tends to infinity.},
under assumptions (A1-A3, A10),
$
\|\fhat - \fstar\hnorm{}  \stackrel{p}{\rightarrow} 0
$
and
$
\Ihat  \stackrel{p}{\rightarrow} \Istar
$.
\end{Theorem}
The condition $\lambdatwo > 0$ is just for 
technical simplicity 
to let $\Sigma_{\Istar,\Istar} + \lambdatwo$ invertible.
The condition $\lambdaone \sqrt{n} \to \infty$ means that $\lambdaone$ does not decrease too quickly. 
The condition \eqref{eq:IRcondsuff} corresponds to an infinite-dimensional extension of the elastic-net
`irrepresentable' condition.  
In the paper of \cite{JMLR:ZhaoYu:2006}, 
the irrepresentable condition 
was derived as a necessary and sufficient condition
for the sign consistency of $\ell_1$ regularization when the number of parameters is finite.
Its elastic-net version was derived in \cite{JRSS:YuanLin:2007},
and it was extended to a situation 
where the number of parameters diverges as $n$ increases
\citep{StatSinica:Jia+Yu:2010}.

We also have a necessary condition for consistency.
\begin{Theorem}
\label{th:NecconsistencyElastMKL}
If $\|\fhat - \fstar\hnorm{} \pto 0$ and 
$\Ihat \pto \Istar$, then 
under assumptions (A1-A3, A10), there exist sequences $\lambdaone,\lambdatwo \to 0$ such that 
\begin{align}
\textstyle 
\limsup_n \left\|\Sigma_{m, \Istar} (\Sigma_{\Istar,\Istar} + \lambdatwo)^{-1}\left(D + 2 \frac{\lambdatwo}{\lambdaone}\right)f^*_{\Istar} \right\hnorm{m}  
\leq 1,~~~~~(\forall m \in J = \Istar^c ).
\label{eq:IRcondness}
\end{align}
Moreover, such $\lambdaone$ satisfies $\lambdaone \sqrt{n} \to \infty$.
\end{Theorem}

The sufficient condition \eqref{eq:IRcondsuff} contains the strict inequality (`$<$'),
while similar conditions for ordinary (non-block) $\ell_1$ regularization
or ordinary (non-block) elastic-net regularization contain
the weak inequality (`$\leq$'). 
The strict inequality appears because 
each block contains multiple variables
in group lasso and MKL \citep{JMLR:BachConsistency:2008}.

%
The condition $\lambdaone \sqrt{n} \to \infty$ is necessary to 
impose the RKHS-norm convergence $\|\fhat - \fstar\hnorm{} \pto 0$.
Roughly speaking, this means that the block-$\ell_1$ regularization term 
should be stronger than the noise level to suppress fluctuations by noise. 

It is worth noting that the conditions \eqref{eq:IRcondsuff} and \eqref{eq:IRcondness} are 
weaker than the condition for {\elloneMKL} presented in \cite{JMLR:BachConsistency:2008};
the {\elloneMKL} irrepresentable condition is\footnote{
Note that in the original paper by \citet{JMLR:BachConsistency:2008}, 
the RHS of \eqref{eq:NSconditionInBach}
is $\sum_{m\in \Istar}\|\fstar_m\hnorm{m}$
because the squared group-$\ell_1$ regularizer $(\sum_{m}\|f_m\hnorm{m})^2$ was used.
We can show that 
the squared formulation is actually equivalent to the non-squared formulation
in the sense that there exists one-to-one correspondence between
the two formulations.
} 
\begin{flalign}
\label{eq:NSconditionInBach}
  \begin{cases}
\mbox{(Sufficient condition)}&
\left\|\Sigma_{m,m}^{1/2}\VCor_{m,\Istar} \VCor_{\Istar,\Istar}^{-1} D \gstar_{\Istar} \right\hnorm{m}  
  <1,~(\forall m \in J),\\
\mbox{(Necessary condition)}&
\left\|\Sigma_{m,m}^{1/2}\VCor_{m,\Istar} \VCor_{\Istar,\Istar}^{-1} D \gstar_{\Istar} \right\hnorm{m}  
\leq 1,~(\forall m \in J).
  \end{cases}
\end{flalign}
This is because the group-$\ell_2$ regularization term 
eases the singularity of the problem. 
Examples that elastic-nets successfully estimate
the true sparsity pattern, while $\ell_1$ regularization
fails in parametric situations can be found in \cite{StatSinica:Jia+Yu:2010}. 

\section{Conclusions}
\label{sec:conclusion}
We provided three novel theoretical results
on the support consistency and convergence rate of {\elasticMKL}.
\begin{enumerate}
\item[(i)] {\ElasticMKL} was shown to be support consistent under a milder condition than {\elloneMKL}. 
\item[(ii)]  A tighter convergence rate 
than existing bounds was derived for the situation where the truth is sparse. 
\item[(iii)] The convergence rates of {\elloneMKL}, {\elasticMKL}, and {\elltwoMKL} 
when the truth is near sparse were elucidated,
and {\elasticMKL} was shown to perform better
when the \emph{decrease rate} $\beta$ is not large,
or the condition of the problem is bad. 
\end{enumerate}
Based on our theoretical findings, we conclude that 
the use of elastic-net regularization is recommended for MKL.

{\ElasticMKL} can be regarded as `intermediate' between {\elloneMKL} and {\elltwoMKL}.
Another popular intermediate variant is 
{\ellpMKL} for $1\le p\le 2$ \citep{NIPS:Marius+etal:2009,UAI:Cortes+etal:2009}.
{\ElasticMKL} and {\ellpMKL} are conceptually similar,
but they have a notable difference:
{\elasticMKL} with $\lambdaone>0$ tends to produce sparse solutions,
while {\ellpMKL} with $1<p\le 2$ always produces dense solutions 
(i.e., all combination coefficients of kernels are non-zero).
Sparsity of {\elasticMKL} would be advantageous 
when the true kernel combination is sparse, as we proved in this paper.
However, when the true kernel combination is non-sparse,
the difference/relation between {\elasticMKL} and {\ellpMKL} is not clear yet.
This needs to be further investigated in the future work.

\appendix
\section{Proofs of the theorems}
For a function $f$ on $\calX \times \Real$, 
we define $P_n f := \frac{1}{n}\sum_{i=1}^n f(x_i,y_i)$
and $P f := \EE_{X,Y}[ f(X,Y)]$.
For a function $f_I \in \calH_I$, 
we define $\|f_I\|_{\ell_1}$ as 
$\|f_I\|_{\ell_1} := \sum_{m\in I} \|f_m\hnorm{m}$ 
and for $f \in \calH$ we write $\|f\|_{\ell_1} := \sum_{m=1}^M \|f_m\hnorm{m}$.
Similarly 
we define $\|f_I\|_{\ell_2}$ as 
$\|f_I\|_{\ell_2}^2 := \sum_{m\in I} \|f_m\hnorm{m}^2$ for $f_I \in \calH_I$
and for $f \in \calH$ we write $\|f\|_{\ell_2}^2 := \sum_{m=1}^M \|f_m\hnorm{m}^2$.
We write $\max\{a,b\}$ as $a \vee b$.


\begin{Lemma}
\label{lem:incoherenceIneq}
For all $I \subseteq \{1,\dots,M\}$, we have
\begin{align}
\| f \|_{\LPi}^2 \geq  (1- \rho(I)^2) \kmin(I) (\sum_{m \in I}\| f_m  \|_{\LPi}^2).
\end{align}
\end{Lemma}
\begin{proof}
For $J = I^c$, we have
\begin{align}
P f^2  &=\|f_I \|_{\LPi}^2 + 2 \langle f_I , f_J \rangle_{\LPi} 
+ \|f_J\|_{\LPi}^2  
\geq 
\|f_I  \|_{\LPi}^2 - 2 \rho(I) \| f_I\|_{\LPi}  \| f_J \|_{\LPi} 
+ \|f_J \|_{\LPi}^2 \notag \\
& \geq 
(1- \rho(I)^2) \| f_I \|_{\LPi}^2
\geq 
(1- \rho(I)^2) \kmin(I) (\sum_{m\in I}\| f_m \|_{\LPi}^2),
\label{eq:firstboundforbasic}
\end{align}
where we used Schwarz's inequality in the last line.
\end{proof}

The following lemma gives an upper bound of $\sum_{m=1}^M\|\fhat \hnorm{m}$
that hold with a high probability. This is an extension of Theorem 1 of \cite{COLT:Koltchinskii:2008}. 
The proof is given in Appendix \ref{sec:proofBasicLemmas}.
\begin{Lemma}
\label{th:boundL1norm}
There exists a constant $F$ depending on only $L$ in {\rm (A1)}
such that, 
if $\lambdaone \geq  F \sqrt{\frac{\log(M n) }{n}} $,
we have,  for $r = \frac{\lambdaone}{\lambdaone \vee \lambdatwo}$, with probability $1 - n^{-1}$,  
\begin{align*}
\sum_{m=1}^M \|\fhat_m \hnorm{m} \leq 
M^{\frac{1-r}{2-r}}
\left( 3 \sum_{m=1}^M \|\fstar_m\hnorm{m} + 3 \sum_{m=1}^M \|\fstar_m\hnorm{m}^2 \right)^{\frac{1}{2-r}}.
\end{align*}
Moreover, if $\lambdatwo \geq  F \sqrt{\frac{\log(M n) }{n}} $ and $\lambdatwo \geq \lambdaone$,
we have 
\begin{align*}
\sum_{m=1}^M \|\fhat_m - \fstar_m \hnorm{m} \leq M \left(3/2 + 2 \max_m \|\fstar_m \hnorm{m} \right).
\end{align*}
\end{Lemma}

The following lemma gives a basic inequality that 
is a start point for the following analyses.  
The proof is given in Appendix \ref{sec:proofBasicLemmas}.
\begin{Lemma}
\label{th:basicineq}
Suppose $\lambdaone \vee \lambdatwo \geq   F \sqrt{\frac{\log(M n) }{n}} $
where $F$ is the constant appeared in Lemma \ref{th:boundL1norm}.
Then there exist constants $\tilde{K}_1$ and $\tilde{K}_2$ depending only on $L$ in {\rm (A1)}, $R$ in {\rm (A4)}, $s$ in ${\rm (A6)}$, $C_2$ in ${\rm (A6)}$ such that 
for all $I \subseteq \{1,\dots,M\}$, 
and for all $t \geq \log \log (R\sqrt{n}) + \log M$,
with probability at least $1-e^{-t} - n^{-1}$, 
\begin{align}
&\frac{1}{2}\|\fhat - \fstar\|_{\LPi}^2  
+  \lambdatwo \sum_{m\in I} \|\fhat_I - \fstar_I \hnorm{m}^2  
+
\lambdatwo \sum_{m \in J} \|\fhat_m\hnorm{m}^2 + \left(\lambdaone - \hat{\gamma}_{n} - \tilde{K}_2 \sqrt{\frac{t}{n}}\right) \sum_{m \in J} \|\fhat_m\hnorm{m}  \notag \\
\leq 
&
\tilde{K}_1 (1+\|\fhat - \fstar\ellone)\Big( 
\sum_{m \in I} \frac{\|\fhat_m - \fstar_m\|_{\LPi}^{1-s} \|\fhat_m - \fstar_m\hnorm{m}^{s} }{\sqrt{n}} \vee \frac{\|\fhat_m - \fstar_m \hnorm{m}}{n^{\frac{1}{1+s}}} 
+ 
\frac{t\|\fhat - \fstar\ellone}{n} \Big) \notag \\
&  \!
+ \!\! \sum_{m\in I} \left(\!\lambdaone \! \frac{\| \gstar_m\hnorm{m}}{\| \fstar_m\hnorm{m}} \!+\! 2\lambdatwo \| \gstar_m\hnorm{m}\!\! + \tilde{K}_2 \sqrt{\frac{t}{n}} \right) 
\!\|\fhat_m - \fstar_m\|_{\LPi} \notag \\
&\!\!+\! \lambdatwo \sum_{m \in J} \|\fstar_m\hnorm{m}^2 \! \!+\! \left(\lambdaone \!\!+\! \hat{\gamma}_{n} + \tilde{K}_2 \sqrt{\frac{t}{n}}\right) \sum_{m \in J} \|\fstar_m\hnorm{m},
\label{eq:ThBoundBasic}
\end{align}
where 
$J= I^c$, 
$\gamma_{n} := \frac{\tilde{K}_1}{\sqrt{n}}$ and $\hat{\gamma}_{n} := \gamma_n(1+\|\fhat - \fstar\|_{\infty}).$ 
\end{Lemma}
The above lemma is derived by {\it peeling device} or {\it localization method}. 
Details of those techniques can be found in, for example, \cite{LocalRademacher,Koltchinskii,IEEEIT:Mendelson:2002,Book:VanDeGeer:EmpiricalProcess}.

\begin{proof}
\textbf{(Theorem \ref{th:SparseRate})}
Since $\lambdaone \geq F \sqrt{\frac{ \log(M n)}{n}}$,
we can assume that the inequality \eqref{eq:ThBoundBasic} is satisfied
with $I = \Istar$.
For notational simplicity, we suppose $I$ denotes $\Istar$ in this proof.
In addition, since $\lambdaone \geq \lambdatwo$, $\|\fhat \|_{\infty} \leq \sum_{m=1}^M \|\fstar \hnorm{m} \leq 3 R$ (with probability $1-n^{-1})$ by Lemma \ref{th:boundL1norm}.
Note that $\|\fstar_m\hnorm{m} = 0$ for all $m \in J = I^c= \Istar^c$,
and  
$\hat{\gamma}_n + \tilde{K}_2 \sqrt{\frac{t}{n}} \leq \max\{\Kconst n^{-\frac{1}{2+s}} + \tilde{K}_2 \sqrt{\frac{t}{n}}, F \sqrt{\frac{ \log(M n)}{n}}\} = \lambdaone$ by taking $\Kconst$ sufficiently large. 
Therefore by the inequality \eqref{eq:ThBoundBasic}, we have
\begin{align}
&\frac{1}{2} \|\fhat - \fstar \|_{\LPi}^2 + \lambdatwo \|\fhat_I - \fstar_I \elltwo^2   
\leq 
K_1 \Big( 
\sum_{m \in I} \frac{\|\fhat_m - \fstar_m\|_{\LPi}^{1-s} \|\fhat_m - \fstar_m\hnorm{m}^{s}}{\sqrt{n}}  
+ 
\frac{t}{n} \Big) \notag \\
&~~~~~~~~~~~~~~~~~~~~~~~~~~~~~~~
+   \sum_{m\in I} \left(\lambdaone\frac{\|\gstar_m\hnorm{m}}{\| \fstar_m\hnorm{m}} + 2\lambdatwo \|\gstar_m\hnorm{m} + \tilde{K}_2 \sqrt{\frac{t}{n}} \right) \|\fhat_m - \fstar_m\|_{\LPi},
\label{eq:BasicIneqSparse}
\end{align}
where $K_1$ is $\tilde{K}_1 (1+3R)$ (here we omitted the term $\sum_{m\in I} n^{-\frac{1}{1+s}} \|\fhat_m - \fstar_m \hnorm{m}$ for simplicity. 
One can show that that term is negligible).

By H{\"o}lder's inequality, the first term in the RHS of the above inequality can be bounded as 
\begin{align*}
K_1 \sum_{m \in I} \frac{\|\fhat_m - \fstar_m\|_{\LPi}^{1-s} \|\fhat_m - \fstar_m\hnorm{m}^{s}}{\sqrt{n}}
&\leq K_1  \frac{(\sum_{m\in I}\|\fhat_m - \fstar_m\|_{\LPi})^{1-s}( \|\fhat_I - \fstar_I \ellone)^{s}}{\sqrt{n}} \\
& \leq \sqrt{d} K_1  \frac{(\sum_{m\in I}\|\fhat_m - \fstar_m\|_{\LPi}^2)^{\frac{1-s}{2}}( \|\fhat_I - \fstar_I \elltwo^2)^{\frac{s}{2}}}{\sqrt{n}}. 
\end{align*}
Applying Young's inequality, the last term 
can be bounded by
\begin{align}
&
 \frac{ K_1 ( \lambdatwo/2)^{-\frac{s}{2}}\sqrt{d} }{\sqrt{n}} (\sum_{m\in I}\|\fhat_m - \fstar_m\|_{\LPi}^2)^{\frac{1-s}{2}} \times ( \lambdatwo/2)^{\frac{s}{2}}( \|\fhat_I - \fstar_I \elltwo^2)^{\frac{s}{2}} \notag \\
\leq &C (n^{- \frac{1}{2}} \sqrt{d} \lambdatwo^{-\frac{s}{2}})^{\frac{2}{2-s}} \left(\sum_{m\in I}\|\fhat_m - \fstar_m\|_{\LPi}^2\right)^{\frac{1-s}{2-s}} 
+ \frac{\lambdatwo}{2}  \|\fhat_I - \fstar_I \elltwo^2  \notag \\
\leq &
C[(1-\rho^2(I))\kmin(I)]^{-1} n^{- 1} d \lambdatwo^{-s} + \frac{(1-\rho^2(I))\kmin(I)}{8} \sum_{m\in I} \|\fhat_m - \fstar_m\|_{\LPi}^2 + \frac{ \lambdatwo}{2} \|\fhat_I - \fstar_I \elltwo^2 \notag \\
\leq &
C n^{- 1} d \lambdatwo^{-s} + \frac{1}{8} \|\fhat - \fstar \|_{\LPi}^2 + \frac{\lambdatwo}{2}  \|\fhat_I - \fstar_I \elltwo^2.
\label{eq:firstTermBoundSparse}  
\end{align} 
where $C$ denotes a constant that is independent of $d$ and $n$ and changes by the contexts, and 
we used Lemma \ref{lem:incoherenceIneq} in the last line.
Similarly, by the inequality of arithmetic and geometric means, we obtain a bound as 
\begin{align}
&\sum_{m\in I} 2 \left(\lambdaone \frac{\| \gstar_m\hnorm{m}}{\| \fstar_m\hnorm{m}} + \lambdatwo  \| \gstar_m\hnorm{m} + \tilde{K}_2 \sqrt{\frac{t}{n}}\right) \|\fhat_m - \fstar_m\|_{\LPi} \notag \\
\leq
& 
C [(1-\rho^2(I))\kmin(I)]^{-1} \sum_{m\in I} \left\{ \left(\frac{\| \gstar_m\hnorm{m}}{\| \fstar_m\hnorm{m}}\right)^2 \lambdaone^2 + \| \gstar_m\hnorm{m}^2 \lambdatwo^2
+ \frac{t}{n} \right\}  \notag \\
&+ \frac{(1-\rho^2(I))\kmin(I)}{8} \sum_{m\in I} \|\fhat_m - \fstar_m\|_{\LPi}^2 \notag \\
\leq
& 
C (d \lambdaone^2 + \lambdatwo^2 + dt/n) + \frac{1}{8} \|\fhat - \fstar\|_{\LPi}^2, 
\label{eq:secondTermBoundSparse}
\end{align}
where we used Lemma \ref{lem:incoherenceIneq} in the last line.
By substituting 
\eqref{eq:firstTermBoundSparse} and \eqref{eq:secondTermBoundSparse} to \eqref{eq:BasicIneqSparse},
we have 
\begin{align}
\frac{1}{4} \|\fhat - \fstar\|_{\LPi}^2 
\leq C \left( d n^{- 1} \lambdatwo^{-s} +  d \lambdaone^2 + \lambdatwo^2 + \frac{(d+1)t}{n}\right).
\end{align}
The minimum of the RHS with respect to $\lambdaone,\lambdatwo$ under the constraint $\lambdaone \geq \lambdatwo$ is achieved 
by $\lambdaone = \max\{\Kconst n^{-\frac{1}{2+s}} + \tilde{K}_2 \sqrt{\frac{t}{n}},F \sqrt{\frac{ \log(M n)}{n}}\}, \lambdatwo= \Kconst n^{-\frac{1}{2+s}}$ up to constants. 
Thus we have the first assertion~\eqref{eq:boundForL2Sparse}.

Next we show the second assertion \eqref{eq:boundForL2Sparse2}. 
By H{\"o}lder's inequality and Young's inequality, we have
\begin{align}
&K_1 \sum_{m \in I} \frac{\|\fhat_m - \fstar_m\|_{\LPi}^{1-s} \|\fhat_m - \fstar_m\hnorm{m}^{s}}{\sqrt{n}}
\leq K_1  \frac{(\sum_{m\in I}\|\fhat_m - \fstar_m\|_{\LPi})^{1-s}( \|\fhat_I - \fstar_I \ellone)^{s}}{\sqrt{n}} \notag \\
& \leq  C  \tilde{\lambda}^{-\frac{s}{1-s}} n^{-\frac{1}{2(1-s)}} \textstyle\sum_{m\in I}\|\fhat_m - \fstar_m\|_{\LPi} + \frac{\tilde{\lambda}}{2} \|\fhat_I - \fstar_I \ellone \notag \\
& \leq  C d  \tilde{\lambda}^{-\frac{2s}{1-s}} n^{-\frac{1}{1-s}} + \frac{1}{8}\textstyle \|\fhat - \fstar\|_{\LPi}^2 + \frac{\tilde{\lambda}}{2} (\|\fhat_I \ellone + \| \fstar_I \ellone),
\label{eq:firstTermBoundSparse2}
\end{align}
where $\tilde{\lambda} >0$ is an arbitrary positive real.
By substituting 
\eqref{eq:firstTermBoundSparse2} and \eqref{eq:secondTermBoundSparse} to \eqref{eq:BasicIneqSparse},
we have 
\begin{align}
&\frac{1}{4} \|\fhat - \fstar \|_{\LPi}^2 \leq 
C \Big( 
  d \tilde{\lambda}^{-\frac{2s}{1-s}} n^{-\frac{1}{1-s}}   + \tilde{\lambda} + d\lambdaone^2 +  \lambdatwo^2 + \frac{(d+1)t}{n} \Big). \notag 
\end{align}
This is minimized by $\tilde{\lambda} = C d^{\frac{1-s}{1+s}}n^{-\frac{1}{1+s}}$, 
$\lambdaone = (\frac{2\tilde{K}_1 (1+3R)}{\sqrt{n}} + \tilde{K}_2\sqrt{\frac{t}{n}})\vee  F \sqrt{\frac{\log(M n) }{n}}  \geq (2\hat{\gamma}_n + \tilde{K}_2\sqrt{\frac{t}{n}})\vee F \sqrt{\frac{\log(M n) }{n}}$, and $\lambdatwo\leq \lambdaone$. Thus we obtain the assertion.
\end{proof}

\begin{proof}
\textbf{(Theorem \ref{th:NearSparse})}
Let $\Id := \{1,\dots,d\}$ and $\Jd = \Id^c = \{d+1,\dots,M\}$.
By the assumption (A7), 
we have 
$
 \sum_{m \in \Jd} \|\fstar_m\hnorm{m}^2 
\leq \frac{C_3}{2 \beta-1}d^{1-2 \beta},~~~
\sum_{m \in \Jd} \|\fstar_m\hnorm{m}
\leq 
\frac{C_3}{\beta-1}d^{1- \beta}.
$
Therefore Lemma \ref{th:basicineq} gives 
\begin{align}
&
  \|\fhat - \fstar \|_{\LPi}^2   
+ \lambdatwo  \|\fhat_{\Id} - \fstar_{\Id} \elltwo^2 
+ \lambdatwo  \|\fhat_{\Jd} \elltwo^2
 \notag \\
\leq 
&K_1 \Big( 
\sum_{m \in \Id} \frac{\|\fhat_m - \fstar_m\|_{\LPi}^{1-s} \|\fhat_m - \fstar_m\hnorm{m}^{s}}{\sqrt{n}}  
+ 
\frac{t  \|\fhat - \fstar \ellone }{n} \Big) \notag \\
&
+
K_1\left(\sum_{m=1}^M \|\fhat_m - \fstar_m\hnorm{m}\right) \Big( 
\sum_{m \in \Id} \frac{\|\fhat_m - \fstar_m\|_{\LPi}^{1-s} \|\fhat_m - \fstar_m\hnorm{m}^{s}}{\sqrt{n}}  
+ 
\frac{t  \|\fhat - \fstar \ellone }{n} \Big)
 \notag \\
&  
+  \sum_{m\in \Id} \!\! \left(\!\lambdaone \! \frac{\| \gstar_m\hnorm{m}}{\| \fstar_m\hnorm{m}} \!+\! 2\lambdatwo \| \gstar_m\hnorm{m}\!\! + \tilde{K}_2\sqrt{\frac{t}{n}} \right)   \|\fhat_m - \fstar_m\|_{\LPi} 
\notag \\
&+
C\left(\lambdatwo  d^{1-2 \beta}  
+ 
\left( \lambdaone + \hat{\gamma}_{n} + \sqrt{\frac{t}{n}}\right) d^{1- \beta}\right), 
\label{eq:firstBoundNearSparse}
\end{align}
if $\lambdaone > \hat{\gamma}_n + \tilde{K}_2\sqrt{\frac{t}{n}} $ and $\lambdaone \geq F\sqrt{\frac{log(Mn)}{n}}$.
The second term can be upper bounded as 
\begin{align*}
&K_1\left(\sum_{m=1}^M \|\fhat_m - \fstar_m\hnorm{m}\right) \Big( 
\sum_{m \in \Id} \frac{\|\fhat_m - \fstar_m\|_{\LPi}^{1-s} \|\fhat_m - \fstar_m\hnorm{m}^{s}}{\sqrt{n}}  
+ 
\frac{t  \|\fhat - \fstar \ellone }{n} \Big) \notag \\
\mathop{\leq}^{\text{H\"older}} 
&
K_1\left(\sum_{m=1}^M \|\fhat_m - \fstar_m\hnorm{m}\right) \Bigg\{ 
\frac{(\sum_{m \in \Id} \|\fhat_m - \fstar_m\|_{\LPi})^{1-s} (\sum_{m \in \Id} \|\fhat_m - \fstar_m\hnorm{m})^{s}}{\sqrt{n}}  
+ 
\frac{t  \|\fhat - \fstar \ellone }{n} \Bigg\} \notag \\
=
&
K_1 
\frac{(\sum_{m \in \Id} \|\fhat_m - \fstar_m\|_{\LPi})^{1-s} \left(\sum_{m=1}^M \|\fhat_m - \fstar_m\hnorm{m}\right)(\sum_{m \in \Id} \|\fhat_m - \fstar_m\hnorm{m})^{s}}{\sqrt{n}}  
+ 
\frac{t  \|\fhat - \fstar \ellone^2 }{n} \notag \\
\mathop{\leq}^{\text{Jensen}}
&
K_1 
\frac{
d^{\frac{1-s}{2}}(\sum_{m \in \Id} \|\fhat_m - \fstar_m\|_{\LPi}^2)^{\frac{1-s}{2}} 
M^{\frac{1}{2}} \left(\sum_{m=1}^M \|\fhat_m - \fstar_m\hnorm{m}^2\right)^{\frac{1}{2}} 
d^{\frac{s}{2}} (\sum_{m \in \Id} \|\fhat_m - \fstar_m\hnorm{m}^2)^{\frac{s}{2}}}{\sqrt{n}}  
\\
&+ 
\frac{t  \|\fhat - \fstar \ellone^2 }{n} \notag \\
\mathop{\leq}^{\text{Lemma \ref{lem:incoherenceIneq}}}
&
K_1 
\frac{
\{(1-\rho(I_d)^2)\kappa(\Id)\}^{-\frac{1-s}{2}} (\|\fhat - \fstar \|_{\LPi}^2)^{\frac{1-s}{2}} 
d^{\frac{1}{2}} M^{\frac{1}{2}} \|\fhat - \fstar \elltwo^{1+s} 
}{\sqrt{n}}  
+ 
\frac{t  \|\fhat - \fstar \ellone^2 }{n} \notag \\
\mathop{\leq}^{\text{Young}}
&
\frac{\|\fhat - \fstar \|_{\LPi}^2}{2} + 
C
\frac{
\{(1-\rho(I_d)^2)\kappa(\Id)\}^{-\frac{1-s}{1+s}}  
d^{\frac{1}{1+s}} M^{\frac{1}{1+s}} \|\fhat - \fstar \elltwo^{2} 
}{n^{\frac{1}{1+s}}}  
+ 
\frac{t  \|\fhat - \fstar \ellone^2 }{n} \notag \\
\mathop{\leq}^{\text{(A8)}}
&
\frac{\|\fhat - \fstar \|_{\LPi}^2}{2} + 
C
\frac{
d^{\frac{b(1-s)+ 1}{1+s}}  
M^{\frac{1}{1+s}} }{n^{\frac{1}{1+s}}} \|\fhat - \fstar \elltwo^{2} 
+ 
\frac{t  \|\fhat - \fstar \ellone^2 }{n}.
\end{align*}
We will see that we may assume $C \frac{d^{\frac{b(1-s)+ 1}{1+s}}M^{\frac{1}{1+s}} }{n^{\frac{1}{1+s}}} \leq \frac{\lambdatwo}{4}$. Thus 
the second term in the RHS of the above inequality can be upper bounded as 
\begin{align}
C\frac{d^{\frac{b(1-s)+ 1}{1+s}}M^{\frac{1}{1+s}} }{n^{\frac{1}{1+s}}} \|\fhat - \fstar \elltwo^{2} 
\leq 
\frac{\lambdatwo}{4} \|\fhat - \fstar \elltwo^{2} 
\leq 
\frac{\lambdatwo}{4} \left( \|\fhat_{\Id} - \fstar_{\Id} \elltwo^2 + 2 \|\fhat_{\Jd}\elltwo^2 + 2\|\fstar_{\Jd} \elltwo^2 \right) \notag \\
\leq 
\frac{\lambdatwo}{2} \left( \|\fhat_{\Id} - \fstar_{\Id} \elltwo^2 +  \|\fhat_{\Jd}\elltwo^2 + \|\fstar_{\Jd} \elltwo^2 \right). 
\label{eq:lambdatwolowerbound}
\end{align}
Moreover Lemma \ref{th:boundL1norm} gives 
$\frac{ \|\fhat - \fstar \ellone}{n} \leq \frac{C \sqrt{R M} }{n} \leq C \lambdatwo^2$
and $\frac{\|\fhat - \fstar \ellone^2}{n} \leq \frac{ CRM}{n} \leq C R \lambdatwo^2. $
Therefore \eqref{eq:firstBoundNearSparse} becomes 
\begin{align*}
&  \frac{1}{2} \|\fhat - \fstar \|_{\LPi}^2   
+ \frac{\lambdatwo}{2}  \|\fhat_{\Id} - \fstar_{\Id} \elltwo^2 
+ \frac{\lambdatwo}{2}  \|\fhat_{\Jd} \elltwo^2
  \notag \\
\leq 
&C \Big( 
\sum_{m \in \Id} \frac{\|\fhat_m - \fstar_m\|_{\LPi}^{1-s} \|\fhat_m - \fstar_m\hnorm{m}^{s}}{\sqrt{n}} + 
t \lambdatwo^2 \Big) \notag \\
&+ \!\! \sum_{m\in \Id} \!\! \! \left(\!C_ 1\lambdaone \!+\! 2\lambdatwo \| \gstar_m\hnorm{m}\!\! + \tilde{K}_2\sqrt{\frac{t}{n}} \right) \!\|\fhat_m - \fstar_m\|_{\LPi} 
 \notag \\
&  
+
C
\left(\lambdatwo  d^{1-2 \beta}  
+ 
\left( \lambdaone + \hat{\gamma}_{n} + \sqrt{\frac{t}{n}}\right) d^{1- \beta}\right).
\end{align*}

As in the proof of Theorem \ref{th:SparseRate} (using the relations \eqref{eq:secondTermBoundSparse} and \eqref{eq:firstTermBoundSparse}),
we have
\begin{align*}
&\frac{1}{2} \|\fhat - \fstar \|_{\LPi}^2  
\notag \\
\leq 
& C\Bigg\{ [(1-\rho^2(\Id))\kmin(\Id))]^{-1} \left[d n^{-1}  \lambdatwo^{-s} + d \lambdaone^2 + \lambdatwo^2 + \frac{t}{n}\right] \\
& + \lambdatwo  d^{1-2 \beta} + ( \lambdaone + \hat{\gamma}_{n} + (t/n)^{\frac{1}{2}}) d^{1- \beta}  + t \lambdatwo^2 \Bigg\}.
\end{align*}
Now using the assumption $(1-\rho^2(\Id))\kmin(\Id) \geq C_4 d^{-b}$, 
we have 
\begin{align}
\|\fhat_{\Id} - \fstar_{\Id} \|_{\LPi}^2  
&\leq 
C\Bigg[ d^{1+b} n^{-1}  \lambdatwo^{-s} \!\!+ d^{1+b} \lambdaone^2 \!\!+ d^b \lambdatwo^2
\!\!+ \lambdatwo  d^{ 1-2\beta} \!+ ( \lambdaone + \hat{\gamma}_{n}) d^{ 1- \beta}  + t \lambdatwo^2 \notag \\
& ~~~~~~~~~~~~~~~~~~ + d^{1- \beta}\sqrt{\frac{t}{n}} + \frac{d^{1+b}t}{n}\Bigg].
\label{eq:FinalBoundNearsparse}
\end{align}

Remind that $\hat{\gamma}_{n} = \tilde{K_1}(1 + \|\fhat - \fstar \|_{\infty})/\sqrt{n}$.
Since $\lambdaone \geq F \sqrt{\frac{\log(Mn)}{n}}$, 
Lemma \ref{th:boundL1norm} gives $\|\fhat - \fstar \|_{\infty} \leq \sqrt{M 3 R} +R \leq c \sqrt{M}$ with probability $1-n^{-1}$ for some constant $c > 0$.
Therefore $\hat{\gamma}_n \leq c \sqrt{M/n}$.
The values of $\lambdaone$, $\lambdatwo$ presented in the statement is achieved by minimizing 
the RHS of \Eqref{eq:FinalBoundNearsparse} under the constraint $\lambdaone \geq c \sqrt{M/n}+ \tilde{K}_2 \sqrt{\frac{t}{n}}\geq \hat{\gamma}_n + \tilde{K}_2 \sqrt{\frac{t}{n}}$
and $C \frac{d^{\frac{b(1-s)+ 1}{1+s}}M^{\frac{1}{1+s}} }{n^{\frac{1}{1+s}}} \leq \frac{\lambdatwo}{4}$.

{{\bf i)} Suppose $n^{-\frac{b + 3 \beta -1}{(2\beta+b)(2+s)-1-s}} > c \sqrt{M/n}$, i.e., $\tau \leq \tau_2$.}
Then the RHS of the above inequality
can be minimized by 
$d = n^{\frac{1}{(2\beta+b)(2+s)-1-s}}$, 
$\lambdatwo = \Kconst n^{-\frac{2\beta + b -1}{(2\beta + b)(2+s)-1-s}}$,
and $\lambdaone =\max\{ \Kconst n^{-\frac{b + 3 \beta -1}{(2\beta+b)(2+s)-1-s}} +\tilde{K}_2 \sqrt{\frac{t}{n}},F \sqrt{\frac{\log(Mn)}{n}} \} $ up to constants independent of $n$,
where the leading terms are 
$d^{1+b} n^{-1}  \lambdatwo^{-s} \!\!+ d^b \lambdatwo^2 \!\!+ \lambdatwo  d^{ 1-2\beta} + \lambdaone d^{ 1- \beta}$.
It should be noted that $\lambdaone$ is greater than 
$\hat{\gamma}_n +\tilde{K}_2 \sqrt{\frac{t}{n}}$ because $n^{-\frac{b + 3 \beta -1}{(2\beta+b)(2+s)-1-s}} > c \sqrt{M/n} \geq \hat{\gamma}_n$, 
therefore \eqref{eq:firstBoundNearSparse} is valid. 
Using $\tau \leq \tau_2$, we can show that $C d^{\frac{b(1-s)+ 1}{1+s}}(M/n)^{\frac{1}{1+s}} \leq \lambdatwo/4$ by setting the constant $\Kconst$ sufficiently large,
hence \eqref{eq:lambdatwolowerbound} is valid.
Moreover, since $M >n^{\frac{1}{(2\beta+b)(2+s)-1-s}} = n^{\tau_1}$, we can take $d$ as 
$d = n^{\frac{1}{(2\beta+b)(2+s)-1-s}} \leq M$. 

{{\bf ii)} Suppose $\tau_2 \leq \tau \leq \tau_3$.} 
Then the RHS of the above inequality
can be minimized by 
$d = (M^{2+s}n^{2-s})^{\frac{1}{2\{(2+s)(b+\beta)-s\}}}$, $\lambdatwo = \Kconst (Mn^{- \{2(b+\beta) -1\}})^{\frac{1}{2\{(2+s)(b+\beta-1)+2\}}}$, 
and $\lambdaone = \max\left\{c \sqrt{M/n} +\tilde{K}_2 \sqrt{\frac{t}{n}},F \sqrt{\frac{\log(Mn)}{n}} \right\} \geq \hat{\gamma}_n+\tilde{K}_2 \sqrt{\frac{t}{n}}$ up to constants independent of $n$,
where the leading terms are 
$d^{1+b} n^{-1}  \lambdatwo^{-s} \!\!+ d^b \lambdatwo^2 \!\!+ \lambdaone  d^{ 1-\beta}$.
Since $\lambdaone \geq \hat{\gamma}_n +\tilde{K}_2 \sqrt{\frac{t}{n}}$,  \eqref{eq:firstBoundNearSparse} is valid. 
Using $\tau \leq \tau_3$, we can show that $C d^{\frac{b(1-s)+ 1}{1+s}}(M/n)^{\frac{1}{1+s}} \leq \lambdatwo/4$ by setting the constant $\Kconst$ sufficiently large,
hence \eqref{eq:lambdatwolowerbound} is valid.
Moreover, since $\beta \leq \frac{s(b -1)}{2(1-s)}$ and $\tau_2 \leq \tau$, 
we can show that $d \leq M$.

{{\bf iii)} Suppose $\tau_3 \leq \tau \leq \tau_4$.} 
We take $\lambdaone = \max\left\{c \sqrt{M/n} +\tilde{K}_2 \sqrt{\frac{t}{n}},F \sqrt{\frac{\log(Mn)}{n}} \right\}$. 
Then the RHS of the inequality \eqref{eq:FinalBoundNearsparse} is minimized by 
$\lambdatwo = \Kconst \sqrt{d} \lambdaone \sim  \sqrt{dM/n}$ and $d=(\frac{n}{M})^{\frac{1}{2(b+\beta)}}$ up to constants, 
where the leading terms are 
$d^b \lambdatwo^2 + d^{1+b} \lambdaone^2 \!\!+ \lambdaone  d^{ 1-\beta}$.
Note that since $\lambdaone \geq \hat{\gamma}_n+\tilde{K}_2 \sqrt{\frac{t}{n}}$,  \eqref{eq:firstBoundNearSparse} is valid. 
Using $\tau \leq \tau_4$, we can show that $C d^{\frac{b(1-s)+ 1}{1+s}}(M/n)^{\frac{1}{1+s}} \leq \lambdatwo/4$ by setting the constant $\Kconst$ sufficiently large,
hence \eqref{eq:lambdatwolowerbound} is valid.
Moreover, since $\beta \leq \frac{s(b -1)}{2(1-s)}$ and $n^{\tau_3}\leq M$, 
we have $d = (\frac{n}{M})^{\frac{1}{2(b+\beta)}} \leq M$. 

~

In all settings i) to iii), we can show that $\frac{d^{1-\beta}}{\sqrt{n}} \gtrsim \frac{d^{1+b}}{n}$.
Thus the terms regarding $t$ is upper bounded as 
$d^{1-\beta}\sqrt{\frac{t}{n}} + \frac{d^{1+b}t}{n} + t\lambdatwo^2 \lesssim (\frac{d^{1-\beta}}{\sqrt{n}}+\lambdatwo^2)(\sqrt{t} + t)$.
Through a simple calculation $\frac{d^{1-\beta}}{\sqrt{n}}$ is evaluated as 
i) $\frac{d^{1-\beta}}{\sqrt{n}} \simeq n^{-\frac{(2\beta+b)(2+s)-3-s+2\beta}{2\{(2\beta+b)(2+s)-1-s\}}}$,
ii) $\frac{d^{1-\beta}}{\sqrt{n}} \simeq (M^{(2+s)(1-\beta)}n^{-(4\beta+2b+sb-2)})^{\frac{1}{2\{(\beta+b)(2+s)-s\}}}$, and 
iii) $\frac{d^{1-\beta}}{\sqrt{n}} \simeq (M^{\beta-1}n^{1-2\beta-b})^{\frac{1}{2(\beta+b)}}$
respectively.
Thus we obtain the assertion.
\end{proof}

\begin{proof}
\textbf{(Theorem \ref{th:NearSparseL1L2})}

\noindent{\bf (Convergence rate of block-$\ell_1$ MKL)}

Note that since $\lambdaone > \lambdatwo = 0$, we have $\frac{\lambdaone}{\lambdaone \vee \lambdatwo} = 1$.
Therefore Lemma \ref{th:boundL1norm} gives 
$
\sum_{m=1}^M \| \fhat_m \hnorm{m} \leq 3 R
$
with probability $1 - n^{-1}$.
Thus $\hat{\gamma}_n = \gamma_n(1 + \|\fhat - \fstar\|_{\infty}) \leq \gamma_n (1 + \sum_{m=1}^M \|  \fhat_m \hnorm{m}  + \sum_{m=1}^M \| \fstar_m \hnorm{m}) 
\leq  \gamma_n(1+4 R)$.

When $\lambdatwo = 0$ and $\lambdaone > (1+4 R) \gamma_n + \tilde{K}_2 \sqrt{\frac{t}{n}}$,
as in Lemma \ref{th:basicineq} we have 
with probability at least $1 - e^{-t} -n^{-1}$ 
\begin{align}
&\|\fhat \!-\! \fstar\|_{\LPi}^2 
\! + \! \lambdaone \!\! \sum_{m\in I} \|\fhat_m \hnorm{m} 
  \notag \\
\leq 
&
K_1 \Big( 
\sum_{m \in I} \frac{\|\fhat_m - \fstar_m\|_{\LPi}^{1-s} \|\fhat_m - \fstar_m\hnorm{m}^{s}}{\sqrt{n}}  
+ 
\frac{t}{n} \Big) + \lambdaone  \sum_{m\in I} \|\fstar_m\hnorm{m} +  2 \lambdaone  \sum_{m \in J} \|\fstar_m\hnorm{m} \notag \\
&+\tilde{K}_2 \sum_{m\in I} \sqrt{\frac{t}{n}}\|\fstar_m - \fhat_m \|_{\LPi},
\label{eq:ThBoundBasicNonsparseTmp}
\end{align}
for all $t \geq \log \log (R\sqrt{n}) + \log M$.

We lower bound the term $\lambdaone \!\! \sum_{m\in I} ( \|\fhat_m \hnorm{m} - \|\fstar_m \hnorm{m} )$ in the LHS of the above inequality~\eqref{eq:BasicIneqSparse}.
There exists $c_1 > 0$ only depending $R$ such that 
\begin{align}
\|f_m \hnorm{m} &= \sqrt{\|f_m - \fstar_m \hnorm{m}^2 - 2 \langle f_m - \fstar_m, \fstar_m\rangle_{\calH_m} + \|\fstar_m\hnorm{m}^2}  \notag \\
& \geq c_1 \|f_m - \fstar_m \hnorm{m}^2 - 2 \|\fstar_m \hnorm{m}^{-1}|\langle f_m - \fstar_m, \fstar_m\rangle_{\calH_m}| + \|\fstar_m\hnorm{m}
\label{eq:NormLowerBound}
\end{align}
for all $f_m\in \calH_m$ such that $\|f_m\hnorm{m} \leq 3R$ and $m\in \Istar$.
Remind that $\fstar_m = T_m^{1/2} \gstar_m$, then we have 
$\|f_m \hnorm{m}  \geq c_1  \|f_m - \fstar_m \hnorm{m}^2 - 2 \frac{\|\gstar_m\hnorm{m}}{\|\fstar_m\hnorm{m}}  \| f_m - \fstar_m \|_{\LPi}  + \|\fstar_m\hnorm{m}$. 
Since $\max_m \|\fhat_m\hnorm{m} \leq 3R$ are met with probability $1-n^{-1}$, 
\begin{equation*}
\|\fhat_m \hnorm{m}  \geq c_1  \|\fhat_m - \fstar_m \hnorm{m}^2 - 2 \frac{\|\gstar_m\hnorm{m}}{\|\fstar_m\hnorm{m}}  \| \fhat_m - \fstar_m \|_{\LPi}  + \|\fstar_m\hnorm{m}, 
\end{equation*}
with probability $1-n^{-1}$.

Therefore by the inequality \eqref{eq:ThBoundBasicNonsparseTmp},
we have with probability at least $1 - e^{-t} - n^{-1}$ 
\begin{align}
&\|\fhat \!-\! \fstar\|_{\LPi}^2 
\! + \! \lambdaone \!\! \sum_{m\in I}  \!(c_1  \|\fhat_m \!- \!\fstar_m \hnorm{m}^2\!\! - \!\!2 \frac{\|\gstar_m\hnorm{m}}{\|\fstar_m\hnorm{m}} 
   \| \fhat_m \!-\! \fstar_m \|_{\LPi}\!\!  
+ \!\! \|\fstar_m\hnorm{m})
  \notag \\
\leq 
&
K_1 \Big( 
\sum_{m \in I} \frac{\|\fhat_m - \fstar_m\|_{\LPi}^{1-s} \|\fhat_m - \fstar_m\hnorm{m}^{s}}{\sqrt{n}}  
+ 
\frac{t}{n} \Big) + \lambdaone  \sum_{m\in I} \|\fstar_m\hnorm{m} +  2 \lambdaone  \sum_{m \in J} \|\fstar_m\hnorm{m} \notag \\
&+\tilde{K}_2 \sum_{m\in I} \sqrt{\frac{t}{n}}\|\fstar_m - \fhat_m \|_{\LPi},
\label{eq:ThBoundBasicNonsparse}
\end{align}
for all $t \geq \log \log (R\sqrt{n}) + \log M$.
Thus using Young's inequality 
\begin{align*}
 \|\fhat - \fstar \|_{\LPi}^2 
\leq &C\left[ d^{1+b} n^{-1}  \lambdaone^{-s} + d^{1+b} \lambdaone^2 
+ 2 \lambdaone  d^{ 1- \beta}  + \frac{t(1+d^{1+b})}{n}\right].
\end{align*}
The RHS is minimized by $d = n^{\frac{1}{(2+s)(\beta+b)}}$
and $\lambdaone = \max\left\{\Kconst n^{-\frac{1}{2+s}} + \tilde{K}_2 \sqrt{\frac{t}{n}}, F\sqrt{\frac{\log(Mn)}{n}}\right\}$ (up to constants independent of $n$).
Note that since the optimal $\lambdaone$ obtained above satisfies $\lambdaone > (1+4 R)\gamma_n+ \tilde{K}_2 \sqrt{\frac{t}{n}}$ by taking $\Kconst$ sufficiently large, 
the inequality \eqref{eq:ThBoundBasicNonsparse} is valid.
Moreover the condition $M > n^{\tau_5} = n^{\frac{b+1}{(\beta+b)\{b(2+s) + 2\}}}$ in the statement
ensures $d< M$.
Finally we evaluate the terms including $t$, that is, $\frac{t}{n}d^{1+b} + \sqrt{\frac{t}{n}}d^{1-\beta}$.
We can check that $\frac{1}{n}d^{1+b} \lesssim \sqrt{\frac{1}{n}}d^{1-\beta}$.
Therefore those terms are upper bounded as 
$\frac{t}{n}d^{1+b} + \sqrt{\frac{t}{n}}d^{1-\beta} \lesssim \sqrt{\frac{1}{n}}d^{1-\beta}(\sqrt{t} + t) \simeq
n^{-\frac{4\beta+2b-2+s(b+\beta)}{2(2+s)(b+\beta)}}(\sqrt{t} + t)$.
Thus we obtain the assertion.

~~~~~

\noindent 
{\bf (Convergence rate for block-$\ell_{2}$ MKL)}

When $\lambdaone =0$, 
substituting $I_M$ to $I$ in Lemma \ref{th:basicineq},
and using Young's inequality,
as in the proof of Theorem \ref{th:NearSparse}, the convergence rate of block-$\ell_{2}$ MKL can be evaluated as 
\begin{align}
&\|\fhat_{\Id} - \fstar_{\Id} \|_{\LPi}^2  
\leq 
C\left[ M^{1+b} n^{-1}  \lambdatwo^{-s} + M^{b} \lambdatwo^2 
+   t \lambdatwo^2 + \frac{t}{n}M^{1+b} \right],
\label{eq:L2BoundNearsparse}
\end{align}
with probability $1-e^{-t} - n^{-1}$ (note that since $I=\{1,\dots,M\}$ ($I^c=\emptyset$), we don't need the condition $\lambdaone \geq \hat{\gamma}_n + \tilde{K}_2\sqrt{\frac{t}{n}}$).
$\lambdatwo = \Kconst (\frac{M}{n})^{\frac{1}{2+s}}\vee  F \sqrt{\frac{\log(M n) }{n}} $
gives the minimum of the RHS with respect to $\lambdatwo$ up to constants.
Using $\tau \leq \tau_6$, we can show that $M^{\frac{b(1-s)+ 1}{1+s}}(M/n)^{\frac{1}{1+s}} = M^{\frac{b(1-s)+ 2}{1+s}} n^{-\frac{1}{1+s}} \lesssim \lambdatwo$ by setting the constant $\Kconst$ 
sufficiently large,
hence \eqref{eq:lambdatwolowerbound} is valid.

\end{proof}

\section{Proof of Lemmas \ref{th:boundL1norm} and \ref{th:basicineq}}
\label{sec:proofBasicLemmas}
\begin{proof}
\textbf{(Lemma \ref{th:boundL1norm})}
Since $\fhat$ minimizes the empirical risk \eqref{eq:primalElasticMKLnonpara},
we have 
\begin{align}
&\frac{1}{n}\sum_{i=1}^n \left(\sum_{m=1}^M (\fhat_m(x_i)-\fstar_m(x_i)) \right)^2 +
\lambdaone  \|\fhat \ellone + \lambdatwo  \|\fhat \elltwo^2 \notag \\
\leq & 
\frac{2}{n}\sum_{m=1}^M \sum_{i=1}^n  \epsilon_i (\fhat_m(x_i) - \fstar_m(x_i)) +
\lambdaone \|\fstar \ellone + \lambdatwo  \|\fstar\elltwo^2.
\label{eq:BasicIneqBoundProof}
\end{align}
By Proposition \ref{prop:HilbertBernstein} (Bernstein's inequality in Hilbert spaces, see also Theorem 6.14 of \cite{Book:Steinwart:2008} for example),
there exists a universal constant $C$ such that we have
\begin{align}
& \frac{1}{n}\sum_{i=1}^n  \epsilon_i (\fhat_m(x_i) - \fstar_m(x_i))   \leq 
 \left| \frac{1}{n}\sum_{i=1}^n  \epsilon_i k_m(x_i,\cdot) \right| \|\fhat_m - \fstar_m \hnorm{m} \notag \\ 
\leq & 
C L \sqrt{\frac{ \log(M n)}{n}} \|\fhat_m - \fstar_m \hnorm{m}
\leq C L \sqrt{\frac{ \log(M n)}{n}}(\|\fhat_m\hnorm{m} + \|\fstar_m \hnorm{m})
\label{eq:empiricalBound}
\end{align}
for all $m$ with probability at least $1 - n^{-1}$, where we used the assumption $\frac{ \log(M n)}{n} \leq 1$.
If $\lambdaone \geq 4 CL \sqrt{\frac{ \log(M n)}{n}}$, then
we have 
\begin{align}
\lambdaone \|\fhat \ellone + \lambdatwo \|\fhat \elltwo^2   
\leq 
3 (\lambdaone \vee \lambdatwo)  (\|\fstar \ellone + \|\fstar \elltwo^2),
\end{align}
with probability at least $1 - n^{-1}$.
Set $r = \frac{\lambdaone}{\lambdaone \vee \lambdatwo}$, then by Young's inequality 
and Jensen's inequality, 
the LHS of the above inequality \eqref{eq:BasicIneqBoundProof} is lower bounded by 
\begin{align}
\lambdaone \|\fhat \ellone + \lambdatwo \|\fhat \elltwo^2  
&\geq  (\lambdaone \vee \lambdatwo) (\sum_{m=1}^M \|\fhat_m \hnorm{m}^{2-r}) \notag \\
&\geq  M (\lambdaone \vee \lambdatwo) \left(\frac{1}{M}\sum_{m=1}^M \|\fhat_m \hnorm{m}^{2-r}\right) \notag \\
&\geq 
M^{r -1} (\lambdaone \vee \lambdatwo) \|\fhat \ellone^{2-r}.
\end{align}
Therefore we have the first assertion by setting $F=4 CL$. 


The second assertion can be shown as follows:
by the inequality \eqref{eq:BasicIneqBoundProof} we have 
\begin{align}
&M^{-1} \lambdatwo \left( \| \fhat - \fstar \ellone \right)^2  \leq 
\lambdatwo \| \fhat - \fstar \elltwo^2 \notag \\
\leq &
\frac{2}{n}\sum_{m=1}^M \sum_{i=1}^n  \epsilon_i (\fhat_m(x_i) - \fstar_m(x_i)) 
+ \lambdaone \| \fhat - \fstar \ellone 
+ 
2 \lambdatwo \sum_{m=1}^M \langle \fstar_m, \fstar_m - \fhat_m \rangle_{\calH_m} \notag \\
\leq &
\lambdatwo\left(\frac{3}{2} + 2 \max_m \|\fstar_m \hnorm{m}\right)  \| \fhat - \fstar \ellone 
\end{align}
with probability at least $1 - n^{-1}$,
where we used \eqref{eq:empiricalBound}, 
$\lambdatwo \geq  4 CL \sqrt{\frac{ \log(M n)}{n}}$ and 
$\lambdatwo \geq \lambdaone$ in the last inequality.
\end{proof}

\begin{proof}
\textbf{(Lemma \ref{th:basicineq})} 
In what follows, we assume 
$\|\fhat - \fstar \ellone \leq \bar{R}$
where 
$\bar{R} = 4MR $  (the probability of this event is greater than $1-n^{-1}$ by Lemma \ref{th:boundL1norm}).
Since $\fhat$ minimizes the empirical risk 
we have 
\begin{align}
&P_n(\fhat - Y)^2 + \lambdaone \|\fhat\ellone + \lambdatwo \|\fhat \elltwo^2
\leq 
P_n(\fstar - Y)^2 + \lambdaone \|\fstar\ellone + \lambdatwo \|\fstar\elltwo^2 \notag \\
\Rightarrow~
&P(\fhat - \fstar)^2 + \lambdaone \|\fhat_J\ellone + \lambdatwo \|\fhat_J\elltwo^2
\leq 
(P - P_n)((\fstar - \fhat)^2 + 2 (\fhat - \fstar) \epsilon) +  \notag \\
&
~~~~~~~~~~~~~~~~~~~~~~
+ \lambdaone (\|\fstar_I \ellone - \|\fhat_I \ellone)  + \lambdatwo (\|\fstar_I\elltwo^2-\|\fhat_I \elltwo^2 ) 
+ \lambdaone \|\fstar_J \ellone + \lambdatwo \|\fstar_J\elltwo^2. 
\label{eq:firsttrivialbound}
\end{align}

The second term in the RHS of the above inequality \eqref{eq:firsttrivialbound} can be bounded from above  
as 
\begin{align}
(\|\fstar_I \ellone - \|\fhat_I \ellone) 
&\leq \sum_{m\in I} \langle \nabla \|\fstar_m \hnorm{m}, \fhat_m - \fstar_m \rangle_{\calH_m} \notag \\ 
&= \sum_{m\in I} \frac{\langle \gstar_m , T_{m}^{1/2} (\fhat_m - \fstar_m) \rangle_{\calH_m}}{\|\fstar_m\hnorm{m}}
\leq \sum_{m\in I} \frac{\| \gstar_m \hnorm{m}}{\|\fstar_m\hnorm{m}} 
\| \fhat_m - \fstar_m \|_{\LPi},  
\label{eq:secondboundforbasic}
\end{align}
where we used $\fstar_m = T_{m}^{1/2}\gstar_m$ for $m \in I \subseteq \Istar$.
We also have
\begin{align}
\lambdatwo (\|\fstar_I\elltwo^2-\|\fhat_I \elltwo^2 )
&= 
\lambdatwo (\sum_{m\in I} 2 \langle \fstar_m, \fstar_m - \fhat_m \rangle_{\calH_m} 
- \|\fhat_I - \fstar_I \elltwo^2 ) \notag \\
&\leq 
\lambdatwo (\sum_{m\in I} 2 \| \gstar_m \hnorm{m} 
\| \fhat_m - \fstar_m \|_{\LPi}
- \|\fhat_I - \fstar_I \elltwo^2 ).
\label{eq:thirdboundforbasic}
\end{align}
Substituting 
\eqref{eq:secondboundforbasic}
and \eqref{eq:thirdboundforbasic}
to \eqref{eq:firsttrivialbound},
we obtain 
\begin{align}
& \| \fhat - \fstar \|_{\LPi}^2
+ \lambdatwo \|\fhat_I - \fstar_I \elltwo^2 + \lambdaone \|\fhat_J\ellone + \lambdatwo \|\fhat_J\elltwo^2 \notag \\
\leq 
&
(P-P_n)((\fstar - \fhat)^2 + 2 (\fhat - \fstar) \epsilon)
+ 
 \sum_{m\in I} (\lambdaone \frac{\| \gstar_m \hnorm{m}}{\|\fstar_m\hnorm{m}}  + 2 \lambdatwo\| \gstar_m \hnorm{m}) 
\| \fhat_m - \fstar_m \|_{\LPi} \notag \\
&+ \lambdaone \|\fstar_J \ellone + \lambdatwo \|\fstar_J\elltwo^2.
\label{eq:secondbasicineq}
\end{align}

Finally we evaluate the first term 
$(P-P_n)((\fstar - \fhat)^2 + 2 (\fhat - \fstar) \epsilon)$
in the RHS of the above inequality \eqref{eq:secondbasicineq}
by applying Talagrand's concentration inequality \citep{Talagrand1,Talagrand2,BousquetBenett}.
First we decompose $(P-P_n)((\fstar - \fhat)^2 + 2 (\fhat - \fstar) \epsilon)$ as 
\begin{align*}
(P-P_n)((\fstar - \fhat)^2 + 2 (\fhat - \fstar) \epsilon)
= 
\sum_{m=1}^M
(P-P_n)((\fstar - \fhat)(\fstar_m - \fhat_m) + 2 (\fhat_m - \fstar_m) \epsilon),
\end{align*}
and bound each term $(P-P_n)((\fstar - \fhat)(\fstar_m - \fhat_m) + 2 (\fhat_m - \fstar_m) \epsilon)$ in the summation.
Here suppose $f\in \calH$ satisfies $\|f\|_{\infty} \leq \|f \ellone \leq \hat{R}$ for a constant $\hat{R}~(\leq \bar{R})$.
Since $|\epsilon| \leq L$,
we have 
\begin{subequations}
\begin{align}
&| ff_m + 2 f_m \epsilon | \leq 2 (L + \hat{R})|f| \leq 2 (L+\hat{R})  \|f_m \hnorm{m}, \\
& \sqrt{P(ff_m + 2 f_m \epsilon)^2}  
=
\sqrt{P(f^2f_m^2) + 4 P(f_m^2 \epsilon^2)}  
\leq 
\sqrt{\|f\|_{\LPi}^2 \|f_m\|_{\LPi}^2 + 4 L^2 \|f_m\|_{\LPi}^2}  \notag \\
&
\leq 
 \|f \|_{\LPi}\|f_m\|_{\LPi} + 2L \|f_m\|_{\LPi},
\end{align}
\label{eq:talagLtwoNorm}
\end{subequations}
for all $f\in \calH$.
Let $Q_n f := \frac{1}{n} \sum_{i=1}^n \varepsilon_i f(x_i,y_i)$ where $\{\varepsilon_i\}_{i=1}^n \in \{\pm 1\}^n$ is 
the Rademacher random variable, and 
$\Psi_m(\xi_m,\sigma_m)$ be 
$$
\Psi_m(\xi_m,\sigma_m) := 
\EE[\sup\{Q_n (|f_m|) \mid f_m \in \calH_m, \|f_m \hnorm{m} \leq \xi_m, \|f_m\|_{\LPi} \leq \sigma_m \}].
$$
Then one can show that by the spectral assumptions (A5) 
(equivalently the covering number condition) 
\[
\Psi_m(\xi_m,\sigma_m) 
\leq K_s \left( \frac{\sigma_m^{1-s} \xi_m^{s}}{\sqrt{n}} \vee n^{-\frac{1}{1+s}}\xi_m \right)
\]
where $K_s$ is a constant that depends on $s$ and $C_2$ \citep{IEEEIT:Mendelson:2002}.
Let $\Xi_m(\xi_m,\sigma_m):= \{f_m \in \calH_m \mid \|f_m \hnorm{m} \leq \xi_m, \|f_m \|_{\LPi} \leq \sigma_m \}$.   
Now by Rademacher contraction inequality \citep[Theorem 4.12]{Book:Ledoux+Talagrand:1991},
for given $\{\xi_m, \sigma_m \}_{m\in I}$ and $\hat{R}$ we have 
\begin{align}
&\EE[ \sup\{Q_n (f f_m + 2 f_m \epsilon)   \mid  f\in \calH \text{~such that~} f_m \in \Xi_m(\xi_m,\sigma_m),~\|f\ellone \leq \hat{R}\} ] \notag \\
\leq &
2 (L+\hat{R}) \Psi_m(\xi_m,\sigma_m) 
\leq 2 K_s (L+\hat{R})  \left( \frac{\sigma_m^{1-s} \xi_m^{s}}{\sqrt{n}} \vee n^{-\frac{1}{1+s}}\xi_m \right).
\end{align}
Therefore by the symmetrization argument \citep{Book:VanDerVaart:WeakConvergence}, we have  
\begin{align}
&\EE[ \sup\{(P_n - P)(f f_m + 2 f_m \epsilon)   \mid  f\in \calH \text{~such that~}f_m \in \Xi_m(\xi_m,\sigma_m),~\|f\ellone \leq \hat{R}\} ] \notag \\
\leq &4 K_s (L+\hat{R}) \left( \frac{\sigma_m^{1-s} \xi_m^{s}}{\sqrt{n}} \vee n^{-\frac{1}{1+s}}\xi_m \right). 
\label{eq:talagExpect}
\end{align}

By Talagrand's concentration inequality with 
\eqref{eq:talagLtwoNorm} and \eqref{eq:talagExpect}, 
for given $\hat{R}, \bar{\sigma}, \xi_m, \sigma_m$ 
with probability at least $1 - e^{-t}$~$(t > 0)$, we have
\begin{align}
&\sup_{ f \in \calH: \atop  \|f\|_{\LPi}\leq \bar{\sigma},\|f\|_{\infty} \leq \hat{R}, f_m \in \Xi_m(\xi_m,\sigma_m) } (P_n - P)(ff_m + 2 f_m \epsilon)   
\leq \notag \\ &~~~~~~
\textstyle \sqrt{2} \left( 4 K_s (L+\hat{R})  \left( \frac{\sigma_m^{1-s} \xi_m^{s}}{\sqrt{n}} \vee \frac{\xi_m}{n^{\frac{1}{1+s}}} \right)
+ \sqrt{\frac{t}{n}} (\bar{\sigma}\sigma_m + 2L\sigma_m) + 2(L+\hat{R})\xi_m \frac{t}{n} \right). 
\end{align}
where we used the relation \eqref{eq:talagLtwoNorm}.
Our next goal is to derive an uniform version of the above inequality over
\[
\frac{1}{\sqrt{n}}\leq \hat{R} \leq \bar{R},~~\frac{1}{\sqrt{n}} \leq \bar{\sigma} \leq \bar{R},~~~\frac{1}{\sqrt{n}M} \leq \xi_m \leq \bar{R}~~~\text{and}~~~\frac{1}{\sqrt{n}M} \leq \sigma_m \leq \bar{R}. 
\] 
By considering a grid $\{\hat{R}^{(k_1)},\bar{\sigma}^{(k_2)}, \xi_m^{(k_3)},\sigma_m^{(k_4)}  \}_{k_i=0 (i=1,\dots,4)}^{\log_2(M \bar{R} \sqrt{n})}$ 
such that $\hat{R}^{(k)}:= \bar{R} 2^{-k}$, 
$\bar{\sigma}^{(k)}:= \bar{R} 2^{-k}$, $\xi_m^{(k)} := \bar{R} 2^{-k}$ and $\sigma_m^{(k)} := \bar{R} 2^{-k}$,
we have 
with probability at least $1- (\log (M \bar{R} \sqrt{n}))^{4} e^{-t} \geq 1- (\log (4 R M^2 \sqrt{n}))^{4} e^{-t}$
\begin{align*}
(P_n - P)(ff_m + 2 f_m \epsilon)   
\leq 
& 
K(1+\|f\ellone) \left( \frac{ \|f_m\|_{\LPi}^{1-s} \|f_m\hnorm{m}^{s}}{\sqrt{n}} \vee \frac{\|f_m\hnorm{m}}{n^{\frac{1}{1+s}}} 
 + \frac{t\|f_m\hnorm{m}}{n}
\right) \notag \\
&~~~~~~~~~~~~~~~
\!\!+\!\! \sqrt{\frac{2t}{n}}  (\|f\|_{\LPi}\|f_m\|_{\LPi} + 2L\|f_m\|_{\LPi}), 
\end{align*}
for all $f \in \calH$ such that $\|f_m \hnorm{m} \leq \bar{R}$ and $\|f\ellone \leq \bar{R}$, and for all $t>1$, where 
$K = 4(4K_s L \vee 4K_s \vee 2 L \vee 2)$. 
Summing up this bound for $m=1,\dots,M$, then we obtain   
\begin{align*}
(P_n - P)(f^2 + 2 f \epsilon)   
\leq 
& 
K(1+\|f\ellone) \left( \sum_{m=1}^M \frac{ \|f_m\|_{\LPi}^{1-s} \|f_m\hnorm{m}^{s}}{\sqrt{n}} \vee \frac{\|f_m\hnorm{m}}{n^{\frac{1}{1+s}}} 
 + \frac{t  \|f \ellone}{n}
\right) \notag \\
&~~~~~~~~~~~~~~~
\!\!+\!\! \sqrt{\frac{2t}{n}}  
\left(\|f\|_{\LPi} \sum_{m=1}^M \|f_m\|_{\LPi} + 2L\sum_{m=1}^M \|f_m\|_{\LPi}\right),
\end{align*}
uniformly for all $f \in \calH$ such that $\|f_m \hnorm{m} \leq \bar{R}$~($\forall m$) and $\|f\ellone \leq \bar{R}$ 
with probability at least $1- M (\log (4 R M^2 \sqrt{n}))^{4} e^{-t}$.
Here set 
$
\gamma_{n} = \frac{K}{\sqrt{n}}
$ 
and note that 
$
\sqrt{\frac{2t}{n}} \|f\|_{\LPi} \sum_{m=1}^M \|f_m\|_{\LPi}
\leq \frac{1}{2} \|f\|_{\LPi}^2 + \frac{t}{n} (\sum_{m=1}^M \|f_m\|_{\LPi})^2
\leq \frac{1}{2} \|f\|_{\LPi}^2 + \frac{t}{n} (\|f \ellone)^2
$
then we have 
\begin{align}
(P_n - P)(f^2 + 2 f \epsilon)  
\leq 
&
K (1+\|f\ellone)\Bigg[   \sum_{m \in I} \left( \frac{ \|f_m\|_{\LPi}^{1-s} \|f_m\hnorm{m}^{s}}{\sqrt{n}} \vee \frac{\|f_m\hnorm{m}}{n^{\frac{1}{1+s}}} \right) 
+ \frac{2 t \|f\ellone}{n} \Bigg] \notag \\ 
&+ \gamma_{n} (1+\|f\ellone) \|f_J\ellone + \frac{1}{2} \|f\|_{\LPi}^2
+
2\sqrt{2}L \sqrt{\frac{t}{n}}  \sum_{m=1}^M \|f_m\|_{\LPi}.
\label{eq:talagFinal}
\end{align}
for all $f\in \calH$ such that $\|f_m \hnorm{m} \leq \bar{R}$~($\forall m$) and $\|f \ellone \leq \bar{R}$ with probability at least  
$1- M(\log (4 R M^2 \sqrt{n}))^{4} e^{-t}$.  
We will replace $t$ with $t+5\log M + 4 \log \log(R\sqrt{n})$,
then the probability $1- M(\log (4 R \sqrt{n} M^2))^{4} e^{-t}$ can be replaced with $1-e^{-t}$
and we have  $t+5\log M + 4\log \log(R\sqrt{n}) \leq 6 t$ for all $t \geq \log M + \log \log(R\sqrt{n})$. 
On the event where $\|\fhat - \fstar \ellone \leq \bar{R}$ holds, substituting $\fhat - \fstar$ to $f$ in \eqref{eq:talagFinal}
and replacing $K$ appropriately, 
\eqref{eq:secondbasicineq} yields 
%
\begin{align}
&\frac{1}{2}\|\fhat - \fstar\|_{\LPi}^2  
+  \lambdatwo \sum_{m\in I} \|\fhat_I - \fstar_I \hnorm{m}^2  
+
\lambdatwo \sum_{m \in J} \|\fhat_m\hnorm{m}^2 + (\lambdaone - \hat{\gamma}_{n}) \sum_{m \in J} \|\fhat_m\hnorm{m}  \notag \\
\leq 
&
\tilde{K}_1 (1+\|\fhat - \fstar\ellone)\Big( 
\sum_{m \in I} \frac{\|\fhat_m - \fstar_m\|_{\LPi}^{1-s} \|\fhat_m - \fstar_m\hnorm{m}^{s} }{\sqrt{n}} \vee \frac{\|\fhat_m - \fstar_m \hnorm{m}}{n^{\frac{1}{1+s}}} 
+ 
\frac{t\|\fhat - \fstar\ellone}{n} \Big) \notag \\
&  \!
+ \!\! \sum_{m\in I} \left(\!\lambdaone \! \frac{\| \gstar_m\hnorm{m}}{\| \fstar_m\hnorm{m}} \!+\! 2\lambdatwo \| \gstar_m\hnorm{m}\!\! \right) 
\!\|\fhat_m - \fstar_m\|_{\LPi} 
\!\!+\! \lambdatwo \sum_{m \in J} \|\fstar_m\hnorm{m}^2 \! \!+\! (\lambdaone \!\!+\! \hat{\gamma}_{n}) \sum_{m \in J} \|\fstar_m\hnorm{m} \notag \\
&+ \tilde{K}_2 \sqrt{\frac{t}{n}}\sum_{m=1}^M \|\fhat_m - \fstar_m\|_{\LPi},
\label{eq:tmpFinalBasicIneq}
\end{align}
where $\tilde{K}_1$ and $\tilde{K}_2$ are constants and $\hat{\gamma}_{n} = \gamma_{n} (1+\|f\ellone)$.
Finally since 
$\tilde{K}_2 \sqrt{\frac{t}{n}}\sum_{m=1}^M \|\fhat_m - \fstar_m\|_{\LPi} 
= \tilde{K}_2 \sqrt{\frac{t}{n}}(\sum_{m\in I}\|\fhat_m - \fstar_m\|_{\LPi} + \sum_{m\in J} \|\fhat_m\|_{\LPi} + \sum_{m\in J}\|\fstar_m\|_{\LPi})
\leq 
\tilde{K}_2 \sqrt{\frac{t}{n}}(\sum_{m\in I}\|\fhat_m - \fstar_m\|_{\LPi} + \sum_{m\in J} \|\fhat_m\hnorm{m} + \sum_{m\in J}\|\fstar_m\hnorm{m})$, 
\eqref{eq:tmpFinalBasicIneq} becomes
\begin{align}
&\frac{1}{2}\|\fhat - \fstar\|_{\LPi}^2  
+  \lambdatwo \sum_{m\in I} \|\fhat_I - \fstar_I \hnorm{m}^2  
+
\lambdatwo \sum_{m \in J} \|\fhat_m\hnorm{m}^2 + \left(\lambdaone - \hat{\gamma}_{n} - \tilde{K}_2 \sqrt{\frac{t}{n}}\right) \sum_{m \in J} \|\fhat_m\hnorm{m}  \notag \\
\leq 
&
\tilde{K}_1 (1+\|\fhat - \fstar\ellone)\Big( 
\sum_{m \in I} \frac{\|\fhat_m - \fstar_m\|_{\LPi}^{1-s} \|\fhat_m - \fstar_m\hnorm{m}^{s} }{\sqrt{n}} \vee \frac{\|\fhat_m - \fstar_m \hnorm{m}}{n^{\frac{1}{1+s}}} 
+ 
\frac{t\|\fhat - \fstar\ellone}{n} \Big) \notag \\
&  \!
+ \!\! \sum_{m\in I} \left(\!\lambdaone \! \frac{\| \gstar_m\hnorm{m}}{\| \fstar_m\hnorm{m}} \!+\! 2\lambdatwo \| \gstar_m\hnorm{m}\!\! + \tilde{K}_2 \sqrt{\frac{t}{n}} \right) 
\!\|\fhat_m - \fstar_m\|_{\LPi} \notag \\
&\!\!+\! \lambdatwo \sum_{m \in J} \|\fstar_m\hnorm{m}^2 \! \!+\! \left(\lambdaone \!\!+\! \hat{\gamma}_{n} + \tilde{K}_2 \sqrt{\frac{t}{n}}\right) \sum_{m \in J} \|\fstar_m\hnorm{m},
\end{align}
which yields the assertion.

\end{proof}

\section{Proof of Theorems \ref{th:SufconsistencyElastMKL} and \ref{th:NecconsistencyElastMKL}}

We write the operator norm of $S_{I,J}:\calH_{J} \to \calH_{I}$ as $\|S_{I,J}\|_{\calH_I,\calH_{J}} := \sup\limits_{g_J \in \calH_J, g_J\neq 0} 
\frac{\| S_{I,J} g_J\hnorm{I}}{\|g_J\hnorm{J}}$.

\begin{Definition}
For all $1\leq m,m' \leq M$, we define the empirical (non centered) cross covariance operator $\hSigma_{m,m'}$ as follows:
\begin{equation}
\langle f_m, \hSigma_{m,m'}g_{m'}\rangle_{\calH_m} := \frac{1}{n} \sum_{i=1}^n f_m(x_i) g_{m'}(x_i),
\end{equation}
where $f_m \in \calH_m, g_{m'} \in \calH_{m'}$. 
Analogous to the joint covariance operator $\Sigma$, we define 
the joint empirical cross covariance operator $\hSigma:\calH \to \calH$ as 
$(\hSigma h)_{m} = \sum_{l=1}^M \hSigma_{m,l}h_l$.
We denote by $\hSigma_{m,\epsilon}$ the element of $\calH_m$ such that  
\[
\langle f_m, \hSigma_{m,\epsilon} \rangle_{\calH_m} := \frac{1}{n} \sum_{i=1}^n \epsilon_i f_m(x_i).
\]
\end{Definition}
Let $\bar{R}$ be a constant such that 
$4 (\sum_{m=1}^M \|\fstar_m \hnorm{m} + \sum_{m=1}^M \|\fstar_m \hnorm{m})\leq \bar{R} $.
We denote by $F_n$ the objective function of elastic-net MKL  
\[
F_n(f) := 
\frac{1}{n}\sum_{i=1}^n (f(x_i) -y_i)^2 + \lambdaone \sum_{m=1}^M \|f_m\hnorm{m} + \lambdatwo \sum_{m=1}^M \|f_m\hnorm{m}^2.
\]

\begin{proof} \textbf{(Theorem \ref{th:SufconsistencyElastMKL})}
Let $\ftil \in \oplus_{m\in \Istar}\calH_m$ be the minimizer of $\tilde{F}_n$:
\begin{align*}
&\ftil := \mathop{\arg \min}_{f \in \calH_{\Istar}} \tilde{F}_n(f), \\
\text{where}~~&\tilde{F}_n(f) := 
\frac{1}{n}\sum_{i=1}^n (f(x_i) -y_i)^2 + \lambdaone \sum_{m\in \Istar} \|f_m\hnorm{m} + \lambdatwo \sum_{m\in \Istar} \|f_m\hnorm{m}^2.
\end{align*}

\noindent {\it (Step 1)} We first show that $\ftil \stackrel{p}{\to} \fstar$ with respect to the RKHS norm.
Since $\lambdaone \sqrt{n} \to \infty$, 
as in the proof of Lemma \ref{th:boundL1norm}, the probability of 
$\sum_{m=1}^M \|\fhat_m - \fstar_m \hnorm{m} \leq \sqrt{M} \bar{R}$ goes to 1
(this can be checked as follows: by replacing $\sqrt{\frac{\log(Mn)}{n}}$ in \Eqref{eq:empiricalBound} with $\log(M) \lambdaone$,
then we see that \Eqref{eq:empiricalBound} holds with probability $1-\exp(-\lambdaone^2 n)$).
There exists $c_1$ only depending $\sqrt{M}\bar{R}$ such that 
\begin{align}
&\|f_m \hnorm{m} = \sqrt{\|f_m - \fstar_m \hnorm{m}^2 - 2 \langle f_m - \fstar_m, \fstar_m\rangle_{\calH_m} + \|\fstar_m\hnorm{m}^2}  \notag \\
\geq & c_1 \|f_m - \fstar_m \hnorm{m}^2 - 2 \|\fstar_m \hnorm{m}^{-1}|\langle f_m - \fstar_m, \fstar_m\rangle_{\calH_m}| + \|\fstar_m\hnorm{m}
\label{eq:fmBound}
\end{align}
for all $m\in \Istar$ and  all $f_m\in \calH_m$ such that $\|f_m\hnorm{m} \leq \sqrt{M}\bar{R}$.

Since $\ftil$ minimizes $\tilde{F}_n$, 
if $\sum_{m=1}^M \|\ftil_m - \fstar_m \hnorm{m} \leq \sqrt{M} \bar{R}$ (the probability of which event goes to 1)
we have 
\begin{align}
& \langle \ftil_{\Istar}- \fstar_{\Istar},\hSigma_{\Istar,\Istar} (\ftil_{\Istar}- \fstar_{\Istar}) \rangle_{\calH_{\Istar}} + c_1 \lambdaone  \sum_{m\in \Istar} \|\ftil_m - \fstar_m \hnorm{m}^2 
+ \lambdatwo\sum_{m \in \Istar} \|\ftil_m - \fstar_m \hnorm{m}^2 \notag \\
\leq &
2 \langle \hSigma_{\Istar, \epsilon}, \ftil - \fstar \rangle_{\calH_{\Istar}} 
 + 2  \sum_{m\in \Istar} \left(\frac{1}{\|\fstar_m\hnorm{m}} \lambdaone + \lambdatwo\right)| \langle \ftil_m - \fstar_m, \fstar_m \rangle_{\calH_m} |,
\label{eq:onebasicineq}
\end{align}
where we used the relation \eqref{eq:fmBound}.
By the assumption $\fstar_m = \Sigma_{m,m}^{1/2} \gstar_m$, 
we have $| \langle \ftil_m - \fstar_m, \fstar_m \rangle_{\calH_m}| \leq \|\gstar_m \hnorm{m} \| \ftil_m - \fstar_m\|_{\LPi} $.
By Lemma \ref{lemm:suppSigmaConv} and Lemma \ref{lemm:noiseConvergeHspace}, we have 
\[
\| \Sigma_{m, m'} - \hSigma_{m,m'} \|_{\calH_m,\calH_{m'}} = O_p(1/\sqrt{n}),~~~\|\hSigma_{\Istar, \epsilon}\hnorm{\Istar} = O_p(1/\sqrt{n}). 
\]
Substituting these inequalities 
to \eqref{eq:onebasicineq},
we have 
\begin{align}
&\|\ftil- \fstar\|_{\LPi}^2  + c_1 \lambdaone  \sum_{m\in \Istar} \|\ftil_m - \fstar_m \hnorm{m}^2 
+ \lambdatwo\sum_{m \in \Istar} \|\ftil_m - \fstar_m \hnorm{m}^2 \notag \\
\leq & O_p\left(\frac{\sum_{m\in \Istar}\|\ftil_m - \fstar_m \hnorm{m}}{\sqrt{n}} + (\lambdaone + \lambdatwo)\sum_{m\in \Istar} \| \ftil_m - \fstar_m\|_{\LPi} \right). 
\label{eq:twobasicineq}
\end{align}
Remind that the (non centered) cross correlation operator is invertible. Thus 
there exists a constant $c$ such that  
\begin{align*}
&\|\ftil- \fstar\|_{\LPi}^2 = \langle \ftil_{\Istar}- \fstar_{\Istar}, \Sigma_{\Istar,\Istar}(\ftil_{\Istar}- \fstar_{\Istar}) \rangle_{\calH} 
= \langle \ftil_{\Istar}- \fstar_{\Istar}, \Diag(\Sigma_{m,m}^{1/2}) V_{\Istar,\Istar} \Diag(\Sigma_{m,m}^{1/2}) (\ftil_{\Istar}- \fstar_{\Istar}) \rangle_{\calH_{\Istar}} \\
\geq& c \sum_{m \in \Istar} \langle \ftil_m- \fstar_m, \Sigma_{m,m}(\ftil_m- \fstar_m) \rangle_{\calH_m} 
= c \sum_{m \in \Istar} \| \ftil_m- \fstar_m \|_{\LPi}^2.
\end{align*}
This and \Eqref{eq:twobasicineq} give that using $ab \leq (a^2 + b^2)/2$
\begin{align}
&\|\ftil- \fstar\|_{\LPi}^2  + c_1 \lambdaone  \sum_{m\in \Istar} \|\ftil_m - \fstar_m \hnorm{m}^2 
+ \lambdatwo\sum_{m \in \Istar} \|\ftil_m - \fstar_m \hnorm{m}^2 \notag \\
& \leq O_p\left(\frac{\sum_{m\in \Istar}\|\ftil_m - \fstar_m \hnorm{m}}{\sqrt{n}} + (\lambdaone + \lambdatwo)\sum_{m\in \Istar} \| \ftil_m - \fstar_m\|_{\LPi} \right) \notag \\
&\leq O_p\left( \frac{1}{n\lambdaone} + (\lambdaone + \lambdatwo )^2 \right) + \frac{c_1}{2} \lambdaone \sum_{m\in \Istar}\|\ftil_m - \fstar_m \hnorm{m}^2 +\frac{c}{2} \sum_{m\in \Istar} \| \ftil_m - \fstar_m\|_{\LPi}^2 \notag \\
&\leq O_p\left( \frac{1}{n \lambdaone} + (\lambdaone + \lambdatwo )^2 \right)+ \frac{c_1}{2} \lambdaone \sum_{m\in \Istar}\|\ftil_m - \fstar_m \hnorm{m}^2  +\frac{1}{2} \|\ftil- \fstar\|_{\LPi}^2. \notag
\end{align}
Therefore we have 
\begin{align*}
&\frac{1}{2} \|\ftil- \fstar\|_{\LPi}^2 + \frac{c_1}{2} \lambdaone  \sum_{m\in \Istar} \|\ftil_m - \fstar_m \hnorm{m}^2 
+ \lambdatwo\sum_{m \in \Istar} \|\ftil_m - \fstar_m \hnorm{m}^2 
\leq O_p\left( \frac{1}{n \lambdaone} + (\lambdaone + \lambdatwo )^2 \right) \\ 
\Rightarrow &\sum_{m\in \Istar} \|\ftil_m - \fstar_m \hnorm{m}^2 
\leq  O_p\left(\frac{1}{(c_1 \lambdaone + \lambdatwo) n \lambdaone}+\frac{(\lambdaone + \lambdatwo)^2}{c_1 \lambdaone + \lambdatwo} \right)
= O_p\left(\frac{1}{ n \lambdaone^2}+(\lambdaone + \lambdatwo) \right). 
\end{align*}
This and $\lambdaone \sqrt{n} \to \infty$ gives $\|\ftil - \fstar_{\Istar}\hnorm{\Istar} \to 0$ in probability.

~

\noindent{\it (Step 2)} Next we show that the probability of $\ftil = \fhat$ goes to 1.
Since $\|\ftil - \fstar_{\Istar}\hnorm{\Istar} \to 0$, we can assume that $\|\ftil_m\hnorm{m} > 0~(m\in \Istar)$  without loss of generality.
We identify $\ftil$ as an element of $\calH$ by setting $\ftil_m = 0$ for $m \in \Jstar$. 
Now we show that $\ftil$ is also the minimizer of $F_n$, that is
$\ftil = \fhat$ , with high probability, hence  
$\Ihat = \Istar$ with high probability.
By the KKT condition, 
the necessary and sufficient condition that $\ftil$ also minimizes $F_n$ is  
\begin{align}
&\| 2 \hSigma_{m,\Istar}(\ftil_{\Istar} - \fstar_{\Istar}) - 2\hSigma_{m, \epsilon} \hnorm{m} \leq 
\lambdaone~~~(\forall m \in \Jstar), 
\label{eq:conditionOneftilopt}
\\
&(2\hSigma_{\Istar,\Istar} + 2 \lambdatwo + \lambdaone D_n) (\ftil_{\Istar} - \fstar_{\Istar}) + \lambdaone D_n \fstar_{\Istar} + 2 \lambdatwo \fstar_{\Istar} -2 \hSigma_{\Istar,\epsilon}= 0,  
\label{eq:conditionTwoftilopt}
\end{align}
where $D_n = \Diag(\|\ftil_m\hnorm{m}^{-1})$.
Note that \eqref{eq:conditionTwoftilopt} is satisfied (with high probability) 
because $\ftil$ is the minimizer of $\tilde{F}_n$ and $\|\ftil_m\hnorm{m} > 0$ for all $m\in \Istar$ (with high probability).
Therefore if the condition \eqref{eq:conditionOneftilopt} holds w.h.p.,
$\ftil = \fhat$ w.h.p..

We will now show the condition \eqref{eq:conditionOneftilopt} holds w.h.p..
Due to \eqref{eq:conditionTwoftilopt}, we have 
\[
\ftil_{\Istar} - \fstar_{\Istar} = -(2\hSigma_{\Istar,\Istar} + 2\lambdatwo + \lambdaone D_n)^{-1} [(\lambdaone D_n + 2\lambdatwo)\fstar_{\Istar}
- 2 \hSigma_{\Istar, \epsilon}].
\]
Therefore the LHS of \eqref{eq:conditionOneftilopt}, 
$\|2\hSigma_{m,\Istar} (\ftil_{\Istar}- \fstar_{\Istar}) - 2\hSigma_{m, \epsilon} \hnorm{m}$, can be evaluated as  
\begin{align}
&\| - 2\hSigma_{m,\Istar}  (2\hSigma_{\Istar,\Istar} + 2\lambdatwo + \lambdaone D_n)^{-1} [(\lambdaone D_n + 2\lambdatwo)\fstar_{\Istar}
- 2 \hSigma_{\Istar, \epsilon}] -2 \hSigma_{m, \epsilon} \hnorm{m} \notag \\
=
&\|2\hSigma_{m,\Istar}  (2\hSigma_{\Istar,\Istar} + 2\lambdatwo + \lambdaone D_n)^{-1} (\lambdaone D_n + 2\lambdatwo)\fstar_{\Istar}  \notag \\
&-  2\hSigma_{m,\Istar}  (2\hSigma_{\Istar,\Istar} + 2\lambdatwo + \lambdaone D_n)^{-1}  2 \hSigma_{\Istar, \epsilon} +2 \hSigma_{m, \epsilon} \hnorm{m} \notag\\
\leq
&\|2\hSigma_{m,\Istar}  (2\hSigma_{\Istar,\Istar} + 2\lambdatwo + \lambdaone D_n)^{-1} (\lambdaone D_n + 2\lambdatwo)\fstar_{\Istar} \hnorm{m} \notag \\
&+ \| 2\hSigma_{m,\Istar}  (2\hSigma_{\Istar,\Istar} + 2\lambdatwo + \lambdaone D_n)^{-1}  2 \hSigma_{\Istar, \epsilon} -2 \hSigma_{m, \epsilon} \hnorm{m}.  
\label{eq:TransKKTonIndicesJ}
\end{align}
We evaluate the probabilistic orders of the last two terms. 

\noindent 
(i) (Bounding $B_{n,m} := \| 2\hSigma_{m,\Istar}  (2\hSigma_{\Istar,\Istar} + 2\lambdatwo + \lambdaone D_n)^{-1}  2 \hSigma_{\Istar, \epsilon} -2 \hSigma_{m, \epsilon} \hnorm{m}$)
We show that 
\begin{align*}
&\hSigma_{m,\Istar}  (2\hSigma_{\Istar,\Istar} + 2\lambdatwo + \lambdaone D_n)^{-1}  \hSigma_{\Istar, \epsilon} = O_p\left(\frac{1}{\sqrt{n}}\right). 
\end{align*}
Since $O \preceq 
\begin{pmatrix}
\hSigma_{\Istar,\Istar} & \hSigma_{\Istar,m}  \\
\hSigma_{m,\Istar} & \hSigma_{m,m} 
\end{pmatrix},
$ 
we have  
\begin{align*}
O\preceq
\begin{pmatrix}
\hSigma_{\Istar,\Istar} + \lambdatwo + \lambdaone D_n/2 & \hSigma_{\Istar,m}  \\
\hSigma_{m,\Istar} & \hSigma_{m,m} + \lambdatwo 
\end{pmatrix}
\preceq
\begin{pmatrix}
2 \hSigma_{\Istar,\Istar} + 2 \lambdatwo + \lambdaone D_n & 0  \\
0 & 2 \hSigma_{m,m} + 2 \lambdatwo
\end{pmatrix}.
\end{align*}
The second inequality is due to the fact that for all $(f_{\Istar}, f_m) \in \calH_{\Istar\cup m}$ we have 
\begin{align*}
\left\langle \begin{pmatrix}f_{\Istar} \\ - f_m \end{pmatrix},
\begin{pmatrix}
\hSigma_{\Istar,\Istar} + \lambdatwo + \lambdaone D_n/2 &- \hSigma_{\Istar,m}  \\
-\hSigma_{m,\Istar} & \hSigma_{m,m} + \lambdatwo 
\end{pmatrix}
\begin{pmatrix}f_{\Istar} \\  - f_m \end{pmatrix}
\right\rangle_{\calH_{\Istar\cup m}}
\geq 0
\end{align*}
because of $O \preceq 
\begin{pmatrix}
\hSigma_{\Istar,\Istar} & \hSigma_{\Istar,m}  \\
\hSigma_{m,\Istar} & \hSigma_{m,m} 
\end{pmatrix}.
$ 

Thus we have 
\begin{align}
&
\left\|
\begin{pmatrix}
\hSigma_{\Istar,\Istar} + \lambdatwo + \frac{\lambdaone D_n}{2} & \hSigma_{\Istar,m}  \\
\hSigma_{m,\Istar} & \hSigma_{m,m} + \lambdatwo 
\end{pmatrix}
\begin{pmatrix}
2 \hSigma_{\Istar,\Istar} + 2 \lambdatwo + \lambdaone D_n & 0  \\
0 & 2 \hSigma_{m,m} + 2 \lambdatwo
\end{pmatrix}^{-1}
\begin{pmatrix}\hat{\Sigma}_{\Istar,\epsilon} \\ \hat{\Sigma}_{m,\epsilon} \end{pmatrix}
\right\|_{\calH_{\Istar \cup m}} \notag \\
&\leq 
\left\|
\begin{pmatrix}\hat{\Sigma}_{\Istar,\epsilon} \\ \hat{\Sigma}_{m,\epsilon} \end{pmatrix}
\right\|_{\calH_{\Istar\cup m}} \leq O_p(1/\sqrt{n}).
\label{eq:SigmaIEpsBound}
\end{align}
Here the LHS of the above inequality is equivalent to
\begin{align*}
\left\|
\begin{pmatrix}
* \\
\hSigma_{m,\Istar}(2\hSigma_{\Istar,\Istar} + 2\lambdatwo + \lambdaone D_n)^{-1} \hat{\Sigma}_{\Istar,\epsilon}
+
(\hSigma_{m,m}+\lambdatwo)(2\hSigma_{m,m} + 2\lambdatwo )^{-1} \hat{\Sigma}_{m,\epsilon}
\end{pmatrix}
\right\|_{\calH_{\Istar\cup m}}. 
\end{align*}
Therefore we observe 
\begin{align*}
\left\|
\hSigma_{m,\Istar}(2\hSigma_{\Istar,\Istar} + 2\lambdatwo + \lambdaone D_n)^{-1} \hat{\Sigma}_{\Istar,\epsilon}
+\frac{1}{2} \hat{\Sigma}_{m,\epsilon}\right\|_{\calH_m} = O_p(1/\sqrt{n}).
\end{align*}
Since $\|\hat{\Sigma}_{m,\epsilon}\hnorm{m}=O_p(1/\sqrt{n})$, we also have 
$$
\|\hSigma_{m,\Istar}(2\hSigma_{\Istar,\Istar} + 2\lambdatwo + \lambdaone D_n)^{-1} \hat{\Sigma}_{\Istar,\epsilon}\hnorm{m}=O_p(1/\sqrt{n}).
$$
This and $\|\hat{\Sigma}_{m,\epsilon}\hnorm{m}=O_p(1/\sqrt{n})$ yield 
\begin{align}
B_{n,m} = O_p(1/\sqrt{n}).
\label{eq:Bnconvergencerate}
\end{align}

~~\\

\noindent 
(ii) (Bounding $E_{n,m} := \|2\hSigma_{m,\Istar}  (2\hSigma_{\Istar,\Istar} + 2\lambdatwo + \lambdaone D_n)^{-1} (\lambdaone D_n + 2\lambdatwo)\fstar_{\Istar} \hnorm{m} $)
Note that, due to $\|\ftil - \fstar\hnorm{} \stackrel{p}{\to} 0$, we have $D_n \stackrel{p}{\to} D$,
and we know that $\max_{m,m'} \|\hSigma_{m ,m'} - \Sigma_{m ,m'}\|_{\calH_m, \calH_{m'}} = O_p(\sqrt{\log(M)/n})= O_p(\frac{1}{\sqrt{n}})$ by Lemma \ref{lemm:suppSigmaConv}.
Thus $S_n := (2\Sigma_{\Istar,\Istar} - 2\hSigma_{\Istar,\Istar})/\lambdaone + D - D_n$ satisfies $S_n = o_p(1)$ and thus $D - S_n \succeq D/2$ with high probability. 
Hence
\begin{align}
&2\hSigma_{m,\Istar}  (2\hSigma_{\Istar,\Istar} + 2\lambdatwo + \lambdaone D_n)^{-1} (\lambdaone D_n + 2\lambdatwo)\fstar_{\Istar} \notag \\
=& 
2\Sigma_{m,\Istar}  (2\hSigma_{\Istar,\Istar} + 2\lambdatwo + \lambdaone D_n)^{-1} (\lambdaone D_n + 2\lambdatwo)\fstar_{\Istar} + 
O_p\left(\frac{1}{\sqrt{n}}\right) \notag \\
=& 
2\Sigma_{m,\Istar}  (2\Sigma_{\Istar,\Istar} + 2\lambdatwo + \lambdaone D)^{-1} (\lambdaone D_n + 2\lambdatwo)\fstar_{\Istar} + \notag \\
&
2\Sigma_{m,\Istar}  (2\Sigma_{\Istar,\Istar} + 2\lambdatwo + \lambdaone D)^{-1} 
\lambdaone S_n 
(2\Sigma_{\Istar,\Istar} + 2\lambdatwo + \lambdaone (D - S_n ))^{-1}
(\lambdaone D_n + 2\lambdatwo)\fstar_{\Istar}  \notag \\
&+O_p\left(\frac{1}{\sqrt{n}}\right).
\label{eq:hSigmaDecomposition}
\end{align}
Here we obtain 
\begin{align}
 &\|\Sigma_{m,\Istar}  (2\Sigma_{\Istar,\Istar} + 2\lambdatwo + \lambdaone D)^{-\frac{1}{2}} \|_{\calH_m,\calH_{\Istar}}^2 \notag \\
=&\|\Sigma_{m,\Istar}  (2\Sigma_{\Istar,\Istar} + 2\lambdatwo + \lambdaone D)^{-1} \Sigma_{\Istar,m} \|_{\calH_m,\calH_m} \notag \\
\leq &
\|\Sigma_{m,m}^{\frac{1}{2}} \VCor_{m,\Istar}  (2\VCor_{\Istar,\Istar})^{-1} \VCor_{\Istar,m} \Sigma_{m,m}^{\frac{1}{2}} \|_{\calH_m,\calH_m}
=O_p(1), 
\label{eq:SigmaSigmainvUpperbound}
\end{align}
and due to the fact that  $D - S_n \succeq D/2$ with high probability we have
\begin{align*}
&\|(\Sigma_{\Istar,\Istar} + \lambdatwo + \lambdaone (D - S_n ))^{-\frac{1}{2}}(\lambdaone D_n + 2\lambdatwo)\fstar_{\Istar} \hnorm{\Istar} \\
=
&
\|(\Sigma_{\Istar,\Istar} + \lambdatwo + \lambdaone (D - S_n ))^{-\frac{1}{2}} \Diag(\Sigma_{m,m}^{\frac{1}{2}})  (\lambdaone D_n + 2\lambdatwo) \gstar_{\Istar} \hnorm{\Istar} \\
\leq
&
O_p(\|\VCor_{\Istar,\Istar}^{-1}\|_{\calH_{\Istar},\calH_{\Istar}}^{-\frac{1}{2}} (\lambdaone  + \lambdatwo))
=O_p(\lambdaone  + \lambdatwo).
\end{align*}
Therefore the second term in the RHS of \Eqref{eq:hSigmaDecomposition} is evaluated as 
\begin{align*}
 &\|\Sigma_{m,\Istar}  (2\Sigma_{\Istar,\Istar} + 2\lambdatwo + \lambdaone D)^{-1} 
\lambdaone S_n 
(2\Sigma_{\Istar,\Istar} + 2\lambdatwo + \lambdaone \!(D \!-\! S_n ))^{-1}
(\lambdaone D_n + 2\lambdatwo)\fstar_{\Istar} \hnorm{m}  \notag \\
\leq
&
\|\Sigma_{m,\Istar}  (2\Sigma_{\Istar,\Istar} + 2\lambdatwo + \lambdaone D)^{-\frac{1}{2}} \|_{\calH_m,\calH_{\Istar}} 
\|(2\Sigma_{\Istar,\Istar} + 2\lambdatwo + \lambdaone D)^{-\frac{1}{2}} \|_{\calH_{\Istar},\calH_{\Istar}} 
\lambdaone \| S_n \|_{\calH_{\Istar},\calH_{\Istar}} \times \\
&\|(2\Sigma_{\Istar,\Istar} + 2\lambdatwo + \lambdaone \!(D \!-\! S_n ))^{-\frac{1}{2}}\|_{\calH_{\Istar},\calH_{\Istar}} 
\|(\Sigma_{\Istar,\Istar} + \lambdatwo + \lambdaone \!(D \!-\! S_n ))^{-\frac{1}{2}}(\lambdaone D_n + 2\lambdatwo)\fstar_{\Istar} \hnorm{\Istar} \\
\leq
& O_p(1 \cdot (\lambdaone + \lambdatwo)^{-\frac{1}{2}} \cdot \lambdaone o_p(1) \cdot (\lambdaone + \lambdatwo)^{-\frac{1}{2}} \cdot (\lambdaone  + \lambdatwo)) \\
=&o_p(\lambdaone ).
\end{align*}
Therefore 
this and \Eqref{eq:hSigmaDecomposition} give
\begin{align*}
&2\hSigma_{m,\Istar}  (2\hSigma_{\Istar,\Istar} + 2\lambdatwo + \lambdaone D_n)^{-1} (\lambdaone D_n + 2\lambdatwo)\fstar_{\Istar} \notag \\
=& 
2\Sigma_{m,\Istar}  (2\Sigma_{\Istar,\Istar} + 2\lambdatwo + \lambdaone D)^{-1} (\lambdaone D_n + 2\lambdatwo)\fstar_{\Istar} + o_p(\lambdaone)+O_p\left(\frac{1}{\sqrt{n}}\right) \\
=&2\Sigma_{m,\Istar}  (2\Sigma_{\Istar,\Istar} + 2\lambdatwo + \lambdaone D)^{-1} (\lambdaone D_n + 2\lambdatwo)\fstar_{\Istar} + o_p(\lambdaone).
\end{align*}

Define 
\begin{align*}
&A_n := \Sigma_{m,\Istar}  \left(\Sigma_{\Istar,\Istar} + \lambdatwo + \frac{\lambdaone D}{2}\right)^{-1} \left( D_n + 2 \frac{\lambdatwo}{\lambdaone}\right)\fstar_{\Istar}, \\
&A  :=  \Sigma_{m,\Istar} (\Sigma_{\Istar,\Istar} + \lambdatwo)^{-1}  \left(D + 2 \frac{\lambdatwo}{\lambdaone}\right)f^*_{\Istar}.
\end{align*}
We show $\|A_n - A\hnorm{m} = o_p(1)$.
By the definition, we have 
\begin{align}
A - A_n = & \Sigma_{m,\Istar} (\Sigma_{\Istar,\Istar} + \lambdatwo)^{-1} \frac{\lambdaone D}{2} 
\left(\Sigma_{\Istar,\Istar} + \lambdatwo + \frac{\lambdaone D}{2}\right)^{-1}
\left( D + 2 \frac{\lambdatwo}{\lambdaone}\right)\fstar_{\Istar} \notag \\
&+ \Sigma_{m,\Istar}  \left(\Sigma_{\Istar,\Istar} + \lambdatwo + \frac{\lambdaone D}{2}\right)^{-1} \left( D - D_n\right)\fstar_{\Istar}.
\label{eq:AnAminus}
\end{align}
On the other hand, as in \Eqref{eq:SigmaIEpsBound}, we observe that   
\begin{align}
2 \geq
&
\left\|
\begin{pmatrix}
\Sigma_{\Istar,\Istar} & \Sigma_{\Istar,m} \\
\Sigma_{m,\Istar} & \Sigma_{m,m}
\end{pmatrix}
\begin{pmatrix}
(\Sigma_{\Istar,\Istar} + \lambdatwo)^{-1} & 0 \\
0 & 0
\end{pmatrix}
\right\|_{\calH_{\Istar \cup m},\calH_{\Istar \cup m}} \notag
\\
=
&
\left\|
\begin{pmatrix}
* & * \\
\Sigma_{m,\Istar} (\Sigma_{\Istar,\Istar} + \lambdatwo)^{-1} & 0
\end{pmatrix}
\right\|_{\calH_{\Istar \cup m},\calH_{\Istar \cup m}}
\geq \|\Sigma_{m,\Istar} (\Sigma_{\Istar,\Istar} + \lambdatwo)^{-1}\|_{\calH_{m},\calH_{\Istar}}.
\label{eq:SigmaIIfirstbound}
\end{align}
Moreover, since $\fstar_{m} = \Sigma_{m,m}^{\frac{1}{2}} \gstar_m$ ($\forall m$), we have 
\begin{align}
&\left\|\left(\Sigma_{\Istar,\Istar} + \lambdatwo + \frac{\lambdaone D}{2}\right)^{-1}
\left( D + 2 \frac{\lambdatwo}{\lambdaone}\right)\fstar_{\Istar}\right\hnorm{\Istar} \notag \\
=
&
\left\|\left(\Sigma_{\Istar,\Istar} + \lambdatwo + \frac{\lambdaone D}{2}\right)^{-1}
\Diag(\Sigma_{m,m}^{\frac{1}{2}}) \left( D + 2 \frac{\lambdatwo}{\lambdaone}\right)\gstar_{\Istar}\right\hnorm{\Istar} \notag \\
\leq
&
\left\|\left(\Sigma_{\Istar,\Istar} + \lambdatwo + \frac{\lambdaone D}{2}\right)^{-\frac{1}{2}}\right\|_{\calH_{\Istar},\calH_{\Istar}}
\left\|\left(\Sigma_{\Istar,\Istar} + \lambdatwo + \frac{\lambdaone D}{2}\right)^{-\frac{1}{2}}\Diag(\Sigma_{m,m}^{\frac{1}{2}})\right\|_{\calH_{\Istar},\calH_{\Istar}} \notag \\
&\times \left\|\left( D + 2 \frac{\lambdatwo}{\lambdaone}\right)\gstar_{\Istar}\right\hnorm{\Istar} \notag \\
\leq
&
O_p((\lambdaone + \lambdatwo)^{-\frac{1}{2}} \left\|\VCor_{\Istar,\Istar}^{-\frac{1}{2}}\right\|_{\calH_{\Istar},\calH_{\Istar}} ) \leq
O_p(\lambdaone^{-\frac{1}{2}}).
\label{eq:SigmaIIfsecondbound}
\end{align}
We can also bound the second term of \eqref{eq:AnAminus} as 
\begin{align*}
& \left\|\Sigma_{m,\Istar}  \left(\Sigma_{\Istar,\Istar} + \lambdatwo + \frac{\lambdaone D}{2}\right)^{-1} \left( D - D_n\right)\fstar_{\Istar}\right\|_{\calH_m} \\
\leq &
\left\|\Sigma_{m,\Istar}  \left(\Sigma_{\Istar,\Istar} + \lambdatwo + \frac{\lambdaone D}{2}\right)^{-1} \right \|_{\calH_m,\calH_{\Istar}}  
\left\| \left( D - D_n\right)\fstar_{\Istar}\right\|_{\calH_{\Istar}} \\
\leq &
\left\|\Sigma_{m,\Istar}  \left(\Sigma_{\Istar,\Istar} + \lambdatwo \right)^{-1} \right\|_{\calH_m,\calH_{\Istar}}  
\left\| \left( D - D_n\right)\fstar_{\Istar}\right\|_{\calH_{\Istar}} \\
\leq &
2 \left\| \left( D - D_n\right)\fstar_{\Istar}\right\|_{\calH_{\Istar}}~~~~(\because \text{\Eqref{eq:SigmaIIfirstbound}}) \\
= & o_p(1).
\end{align*}
Therefore 
applying the inequalities \Eqref{eq:SigmaIIfirstbound} and \Eqref{eq:SigmaIIfsecondbound} to \Eqref{eq:AnAminus}, 
we have 
\begin{align}
\|A_n - A\hnorm{m} = O_p(\lambdaone^{\frac{1}{2}}) + o_p(1) = o_p(1).
\label{eq:AnminusAconv}
\end{align}
Hence we have $E_{n,m} = \lambdaone \|A\hnorm{m} + o_p(\lambdaone)$.

~

\noindent (iii) (Combining (i) and (ii))
Due to the above evaluations ((i) and (ii)), we have 
\begin{align*}
&\max_{m \in \Jstar} \left\| 2 \hSigma_{m,I}(\ftil_{\Istar} - \fstar_{\Istar}) - 2\hSigma_{m, \epsilon} \right\hnorm{m} \\
= 
&\max_{m \in J} \lambdaone \left\|\Sigma_{m,\Istar}( \Sigma_{\Istar,\Istar} + \lambdatwo )^{-1} \left(D + 2 \frac{\lambdatwo}{\lambdaone}\right)f^*_{\Istar} \right\hnorm{m} + o_p(\lambdaone)
< \lambdaone (1 - \eta) 
+ o_p(\lambdaone).
\end{align*}
This yields 
\[
P\left(\| 2 \hSigma_{m,\Istar}(\ftil_{\Istar} - \fstar_{\Istar}) - 2\hSigma_{m, \epsilon} \hnorm{m} \geq \lambdaone, 
\forall m \in \Jstar \right) \to 0.
\]
Thus the probability of the condition \eqref{eq:conditionOneftilopt} goes to 1.
\end{proof}

\begin{proof} \textbf{(Theorem \ref{th:NecconsistencyElastMKL})} 
First we prove that $\lambdaone \sqrt{n} \to \infty$ is a necessary condition
for $\Ihat \pto \Istar$.
Assume that $\liminf \lambdaone \sqrt{n} < \infty$.
Then we can take a sub-sequence that converges to a finite value,
therefore by taking the sub-sequence, if necessary, we can assume $\lim \lambdaone \sqrt{n} \to \muone$
without loss of generality.
We will derive a contradiction 
under the conditions of 
$\|\fhat - \fstar\hnorm{} \pto 0$ and $\Ihat \pto \Istar$.
Suppose $\Ihat = \Istar$.

By the KKT condition,
\begin{align}
&0 = 
2(\hSigma_{\Istar,\Istar} \fhat_{\Istar} - \hSigma_{\Istar, \epsilon} - \hSigma_{\Istar,\Istar} \fstar_{\Istar}) + \lambdaone D_n \fhat_{\Istar} + 2\lambdatwo \fhat_{\Istar} \notag \\
\Rightarrow~~&
2(\hSigma_{\Istar,\Istar} + \lambdatwo) (\fstar_{\Istar} - \fhat_{\Istar})  = 
\lambdaone D_n \fstar_{\Istar} + 2\lambdatwo \fstar_{\Istar} - 2\hSigma_{\Istar, \epsilon} \label{eq:KKTeqcondtwo} \\
\Rightarrow~~&
2\sqrt{n}(\Sigma_{\Istar,\Istar} + \lambdatwo) (\fstar_{\Istar} - \fhat_{\Istar})  = 
\sqrt{n} \lambdaone D \fstar_{\Istar} + \sqrt{n} 2\lambdatwo \fstar_{\Istar}
- 2\sqrt{n}\hSigma_{\Istar, \epsilon}
 \notag \\
&~~~~~~~~~~~~~~~~~~+ 
(2\sqrt{n}(\Sigma_{\Istar,\Istar} - \hSigma_{\Istar,\Istar})(\fstar_{\Istar} - \fhat_{\Istar}) +  
\sqrt{n} \lambdaone (D_n - D) \fstar_{\Istar} ) \notag \\
\Rightarrow~~&
2\sqrt{n}(\Sigma_{\Istar,\Istar} + \lambdatwo) (\fstar_{\Istar} - \fhat_{\Istar})  = 
\muone D \fstar_{\Istar} + 
\sqrt{n} 2\lambdatwo \fstar_{\Istar} - 
 2\sqrt{n}\hSigma_{\Istar, \epsilon} + 
o_p(1),
\label{eq:KKTnecessarycond}
\end{align}
where the last inequality is due to $\sqrt{n}\lambdaone \to \muone, \|D_n -D\|_{\calH_{\Istar},\calH_{\Istar}} = o_p(1), \|\fhat - \fstar\hnorm{} = o_p(1)$
and $\|\Sigma_{\Istar,\Istar} - \hSigma_{\Istar,\Istar}\|_{\calH_{\Istar},\calH_{\Istar}} = o_p(1)$.
Moreover since the second equality \eqref{eq:KKTeqcondtwo} indicates that 
$o_p(1) + o_p(\lambdatwo) = \lambdaone D \fstar_{\Istar} + 2\lambdatwo \fstar_{\Istar} + o_p(1)$,
we have $\lambdaone = o_p(1)$ and $\lambdatwo = o_p(1)$.

We now show that the KKT condition under which $\fhat$ satisfying $\Ihat = \Istar$ is optimal with respect to $F_n$ 
is violated with strictly positive probability: 
\begin{align}
\liminf P\left(\exists m \in J,~\|2(\hSigma_{m,\Istar} \fhat_{\Istar} - \hSigma_{m,\Istar}\fstar_{\Istar} - \hSigma_{m,\epsilon})\hnorm{m} > \lambdaone \right) > 0. 
\label{eq:liminfPKKTpositive}
\end{align}
Obviously this indicates that the probability $\Ihat = \Istar$ does not converges to 1, which is a contradiction. 

For all $v_m \in \calH_m$ $(m\in \Jstar)$, there exists $w_{\Istar} \in \calH_{\Istar}$
such that 
\begin{equation}
\Sigma_{\Istar, m}v_m =  (\Sigma_{\Istar,\Istar} + \lambdatwo) w_{\Istar}.
\label{eq:vmequalwI}
\end{equation}
Note that $w_{\Istar}$ is uniformly bounded for all $\lambdatwo \geq 0$ because 
the range of $\Sigma_{\Istar,m}$ is included in the range of $\Sigma_{\Istar,\Istar}$ \citep{TAMS:Baker:1973}
and
there exists $\tilde{w}_{\Istar}$ such that $\Sigma_{\Istar,m}v_m = \Sigma_{\Istar,\Istar} \tilde{w}_{\Istar}$ ($\tilde{w}_{\Istar}$ is independent of $\lambdatwo$), 
hence $\Sigma_{\Istar,\Istar} \tilde{w}_{\Istar} = (\Sigma_{\Istar,\Istar} + \lambdatwo) w_{\Istar}$,
and  
$$\|w_{\Istar}\hnorm{\Istar} \leq \sqrt{\langle \tilde{w}_{\Istar},  \Sigma_{\Istar,\Istar} (\Sigma_{\Istar,\Istar} + \lambdatwo)^{-2} \Sigma_{\Istar,\Istar} \tilde{w}_{\Istar} \rangle_{\calH_{\Istar}}}
\leq \|\tilde{w}_{\Istar}\|_{\calH_{\Istar}}$$
for $\lambdatwo > 0$ and $\|w_{\Istar}\hnorm{\Istar} = \|\tilde{w}_I\hnorm{\Istar}$ for $\lambdatwo =0$.
Let $v_m \in \calH_m$ be any non-zero element such that 
$\Sigma_{m,m}^{1/2} v_m \neq 0$ and $w_{\Istar}$ be satisfying the above equality \eqref{eq:vmequalwI}, then 
\begin{align*}
&\sqrt{n} \langle v_m, \hSigma_{m,\epsilon} + \hSigma_{m,\Istar}\fstar_{\Istar} - \hSigma_{m,\Istar} \fhat_{\Istar} \rangle_{\calH_m} \\
= &
\sqrt{n} \langle v_m, \hSigma_{m, \epsilon}\rangle_{\calH_m}
+ \langle v_m,  \hSigma_{m,\Istar} \sqrt{n}(\fstar_{\Istar} - \fhat_{\Istar}) \rangle_{\calH_m} \\
= &
\sqrt{n} \langle v_m, \hSigma_{m, \epsilon}\rangle_{\calH_m}
+ \langle v_m,  \Sigma_{m, I} \sqrt{n}(\fstar_{\Istar} - \fhat_{\Istar}) \rangle_{\calH_m}
+ 
o_p(1) \\
= &
\sqrt{n} \langle v_m, \hSigma_{m, \epsilon}\rangle_{\calH_m}
+ \langle w_{\Istar}, (\Sigma_{\Istar,\Istar} + \lambdatwo) \sqrt{n}(\fstar_{\Istar} - \fhat_{\Istar}) \rangle_{\calH_m}
+ o_p(1) \\
= &
\sqrt{n} \langle v_m, \hSigma_{m, \epsilon}\rangle_{\calH_m}
-
\sqrt{n} \langle w_{\Istar}, \hSigma_{\Istar, \epsilon} \rangle_{\calH_m}
+
\left \langle w_{\Istar}, \left( \frac{\muone}{2} D  + \sqrt{n} \lambdatwo \right) \fstar_{\Istar} \right\rangle_{\calH_m}
+ o_p(1),
\end{align*}
where we used $\|\hSigma_{m,\Istar} - \Sigma_{m,\Istar}\|_{\calH_m,\calH_{\Istar}} = O_p(1/\sqrt{n})$ and $\|\fstar - \fhat\hnorm{} \pto 0$
in the second equality,
and the relation \eqref{eq:KKTnecessarycond} in the last equality.
We can show that 
$Z_n := \sqrt{n} \langle v_m, \hSigma_{m, \epsilon}\rangle
-
\sqrt{n} \langle w_{\Istar}, \hSigma_{\Istar, \epsilon} \rangle
$
has a positive variance as follows (see also \cite{JMLR:BachConsistency:2008}):
\begin{align*}
\EE[Z_n]  &= 0, \\
\EE[Z_n^2] &\geq 
\sigma^2  \left( \langle v_m, \Sigma_{m, m} v_m \rangle 
-
2 \langle v_m, \Sigma_{m ,\Istar} w_{\Istar} \rangle 
+
\langle w_{\Istar}, \Sigma_{\Istar ,\Istar} w_{\Istar} \rangle  \right) \\
&
= \sigma^2 \left( \langle v_m, \Sigma_{m, m} v_m \rangle 
-
\langle v_m, \Sigma_{m,\Istar} w_{\Istar} \rangle 
+
o_p(1) \right)~~~~~~~(\because \lambdatwo =o_p(1)) \\
&= 
\sigma^2  \langle \Sigma_{m,m}^{1/2} v_m, 
(I_{\calH_m} - \VCor_{m,\Istar} \Ctwo^{-1}_{\Istar,\Istar} \VCor_{\Istar, m}) \Sigma_{m,m}^{1/2}  v_m \rangle + 
o_p(1),
\end{align*}
where $\Ctwo^{-1}_{\Istar,\Istar}  = \Diag(\Sigma_{m ,m}^{1/2}) (\Sigma_{\Istar,\Istar} + \lambdatwo)^{-1}\Diag(\Sigma_{m, m}^{1/2})$
(note that $\Ctwo_{\Istar,\Istar}$ is invertible because $\VCor_{\Istar,\Istar} \preceq \Ctwo_{\Istar,\Istar}$ and $\VCor_{\Istar,\Istar}$ is invertible). 
Now since $\VCor_{\Istar,\Istar} \preceq \Ctwo_{\Istar,\Istar}$ and 
$I_{\calH_m} - \VCor_{m,\Istar} \VCor^{-1}_{\Istar,\Istar}\VCor_{\Istar,m} \succ O$
(this is because $\VCor_{\Istar\cup m, \Istar\cup m} = \begin{pmatrix} \VCor_{\Istar,\Istar} & \VCor_{m,\Istar} \\ \VCor_{\Istar,m} & I_{\calH_m} \end{pmatrix}$ is invertible),
we have 
$I_{\calH_m} - \VCor_{m,\Istar} \Ctwo^{-1}_{\Istar,\Istar} \VCor_{\Istar,m} \succ O$.
Therefore 
by the central limit theorem 
$Z_n$ converges Gaussian random variable with strictly positive variance in distribution.
Thus 
the probability of 
\[
2 |\langle v_m, \hSigma_{m,\epsilon}+\hSigma_{m,\Istar} \fstar_{\Istar} - \hSigma_{m,\Istar} \fhat_{\Istar} \rangle_m | > \lambdaone  \|v_m\hnorm{m}
\]
is asymptotically strictly positive because $\lambdaone \sqrt{n} \to \muone$
(Note that this is true whether $\sqrt{n}\lambdatwo$ converges to finite value or not).
This yields \eqref{eq:liminfPKKTpositive}, i.e.
$\fhat$ does not satisfy $\Ihat = \Istar$ with asymptotically strictly positive probability.


We say Condition A as  
\[
\text{Condition A}:~~~~\lambdaone\sqrt{n} \to \infty. 
\] 


Now that we have proven $\lambdaone \sqrt{n} \to \infty$, 
we are ready to prove the assertion \eqref{eq:IRcondness}. 
Suppose the condition \eqref{eq:IRcondness} is not satisfied
for any sequences $\lambdaone, \lambdatwo \to 0$, that is, 
there exists a constant $\xi > 0$ such that  
\begin{align}
\limsup_{n\to \infty} \left\|\Sigma_{m,\Istar}(\Sigma_{\Istar,\Istar} + \lambdatwo)^{-1} 
\left(D + 2 \frac{\lambdatwo}{\lambdaone}\right)g^*_{\Istar} \right\hnorm{m} > 
(1 + \xi), 
~~(\exists m \in \Jstar),
\label{eq:AntiNecessaryCondition}
\end{align}
for any sequences $\lambdaone, \lambdatwo \to 0$
satisfying Condition A ($\lambdaone \sqrt{n} \to \infty$). 
Fix arbitrary sequences $\lambdaone, \lambdatwo \to 0$ satisfying Condition A. 
If $\Ihat = \Istar$, the KKT condition
\begin{align}
&\| 2 \hSigma_{m,\Istar}(\fhat_{\Istar} - \fstar_{\Istar}) - 2\hSigma_{m, \epsilon} \hnorm{m} \leq \lambdaone~~~(\forall m \in \Jstar), 
\label{eq:conditionOneftiloptdash}
\\
&(2\hSigma_{\Istar,\Istar} + 2 \lambdatwo + \lambdaone D_n) (\ftil_{\Istar} - \fstar_{\Istar}) + \lambdaone D_n \fstar_{\Istar} + 2 \lambdatwo \fstar_{\Istar} -2 \hSigma_{\Istar,\epsilon}= 0,  
\label{eq:conditionTwoftiloptdash}
\end{align}
should be satisfied (see \eqref{eq:conditionOneftilopt} and \eqref{eq:conditionTwoftilopt}). 
We prove that the first inequality \eqref{eq:conditionOneftiloptdash} of the KKT condition is violated with strictly positive probability under the assumptions and 
the condition \eqref{eq:conditionTwoftiloptdash}. 
We have shown that 
(see \eqref{eq:TransKKTonIndicesJ})
\begin{align}
&\lambdaone^{-1}(2 \hSigma_{m,\Istar}(\fhat_{\Istar} - \fstar_{\Istar}) - 2\hSigma_{m, \epsilon}) \notag \\
=&
2\hSigma_{m,\Istar}  (2\hSigma_{\Istar,\Istar} + 2\lambdatwo + \lambdaone D_n)^{-1} ( D_n + 2 \frac{\lambdatwo}{\lambdaone})\fstar_{\Istar}   \notag \\
&-  \frac{2}{\lambdaone}\hSigma_{m,\Istar}  (2\hSigma_{\Istar,\Istar} + 2\lambdatwo + \lambdaone D_n)^{-1}  2 \hSigma_{\Istar, \epsilon} +\frac{2}{\lambdaone}
 \hSigma_{m ,\epsilon}.  
\label{eq:KKTcondModified}
\end{align}
As shown in the proof of Theorem 1, 
the first term can be approximated by
$
\Sigma_{m,\Istar} \left(\Sigma_{\Istar,\Istar} + \lambdatwo\right)^{-1} \left(D + 2 \frac{\lambdatwo}{\lambdaone}\right)f^*_{\Istar},
$
more precisely  \Eqref{eq:AnminusAconv} gives
\begin{align*}
&\left\| \hSigma_{m,\Istar}  \left(\hSigma_{\Istar,\Istar} + \lambdatwo + \frac{\lambdaone D_n}{2}\right)^{-1} 
\!\!\! \left( D_n + 2 \frac{\lambdatwo}{\lambdaone}\right)\fstar_{\Istar}
- 
\Sigma_{m,\Istar} (\Sigma_{\Istar,\Istar} + \lambdatwo)^{-1} \!\left(\! D \! +\! 2 \frac{\lambdatwo}{\lambdaone}\right)g^*_I 
\right\hnorm{m} \\
&\pto 0. 
\end{align*}
Since 
$
\liminf_{n} \left\|\Sigma_{m,\Istar} (\Sigma_{\Istar,\Istar} + \lambdatwo)^{-1} \left(D + 2 \frac{\lambdatwo}{\lambdaone}\right)g^*_{\Istar} \right\hnorm{m}  > (1 + \xi)
$
by the assumption,
we observe that 
\begin{align}
\label{eq:NecessaryCondConstConvergence}
P\left(
\left\| 2\hSigma_{m,\Istar}  (2\hSigma_{\Istar,\Istar} + 2\lambdatwo + \lambdaone D_n)^{-1} \left( D_n + 2 \frac{\lambdatwo}{\lambdaone}\right)\fstar_{\Istar} \right\hnorm{m} > 
(1  + \xi) 
\right) \not\to 0.
\end{align}
Now since $\lambdaone \sqrt{n} \to \infty $,
we have proven that  
\begin{align}
\left\|-  \frac{2}{\lambdaone}\hSigma_{m,\Istar}  (2\hSigma_{\Istar,\Istar} + 2\lambdatwo + \lambdaone D_n)^{-1}  2 \hSigma_{\Istar, \epsilon} +\frac{2}{\lambdaone}
 \hSigma_{m, \epsilon} \right \hnorm{m} 
= O_p(1/(\lambdaone \sqrt{n})) = o_p(1),
\label{eq:NecessaryCondNoiseConvergence}
\end{align}
in the proof of Theorem 1 (\Eqref{eq:Bnconvergencerate}).
Therefore, combining \eqref{eq:KKTcondModified}, \eqref{eq:NecessaryCondConstConvergence} and \eqref{eq:NecessaryCondNoiseConvergence},
we have observed that the KKT condition \eqref{eq:conditionOneftilopt} 
is violated with strictly positive probability if the condition \eqref{eq:AntiNecessaryCondition}
is satisfied.  
This yields the irrepresenter condition \eqref{eq:IRcondness} 
is a necessary condition for the consistency of elastic-net MKL. 
\end{proof}

\begin{Lemma}
\label{lemm:suppSigmaConv}
If $\sup_{X} k_m(X,X) \leq 1$ and $\sup_{X} k_{m'}(X,X) \leq 1$, then 
\begin{align}
P(\|\hSigma_{m, m'} - \Sigma_{m, m'} \|_{\calH_m,\calH_m'} \geq 
\EE[\|\hSigma_{m, m'} - \Sigma_{m ,m'} \|_{\calH_m,\calH_m'}] + \varepsilon)
\leq \exp( - n \varepsilon^2 / 2).
\label{eq:SigmaConvFirstAss}
\end{align}
In particular, 
\begin{equation}
P\left( \|\hSigma_{m, m'}- \Sigma_{m,m'} \|_{\calH_m,\calH_m'} \geq \sqrt{\frac{1}{n}} + \epsilon \right) 
\leq \exp( - n \varepsilon^2 / 2).
\label{eq:SigmaConvSecondAss}
\end{equation}
\end{Lemma}
\begin{proof}
We use McDiarmid's inequality \citep{Book:ProbTheoryDevroye:1996}. 
By definition 
\[
\langle g, \hSigma_{m m'}f \rangle = \frac{1}{n} \sum_{i=1}^n 
\left \langle g, k_m(\cdot, x_i)  \right\rangle_m     
\left \langle f, k_{m'}(\cdot, x_i) \right\rangle_{m'}.    
\]
We denote by $\tSigma_{m ,m'}$ the empirical cross covariance operator with 
$n$ samples $(x_1,\dots,x_{j-1},\tilde{x}_j, x_{j+1},\dots,x_{n})$
where the $j$-th sample $x_j$ is replaced by $\tilde{x}_j$ 
independently distributed by the same distribution as $x_j$'s.

By the triangular inequality,
we have
\[
\|\hSigma_{m ,m'} - \Sigma_{m ,m'} \|_{\calH_m,\calH_{m'}}
-
\|\tSigma_{m ,m'} - \Sigma_{m ,m'} \|_{\calH_m,\calH_{m'}}
\leq
\|\hSigma_{m ,m'} - \tSigma_{m ,m'} \|_{\calH_m,\calH_{m'}}.
\]
Now the RHS can be evaluated as follows:
\begin{align}
&\| \hSigma_{m ,m'} - \tSigma_{m ,m'} \|_{\calH_m,\calH_{m'}} \notag \\
=&
\left\| \frac{1}{n}(k_m(\cdot,x_j)k_{m'}(x_j,\cdot) - k_m(\cdot,\tilde{x}_j)k_{m'}(\tilde{x}_j,\cdot))\right\|_{\calH_m,\calH_{m'}}. 
\label{eq:McDifferenceBasic}
\end{align}
The RHS of \eqref{eq:McDifferenceBasic} can be further evaluated as 
\begin{align}
&\|\frac{1}{n}(k_m(\cdot,x_j)k_{m'}(x_j,\cdot) - k_m(\cdot,\tilde{x}_j)k_{m'}(\tilde{x}_j,\cdot))\|_{\calH_m,\calH_{m'}} \notag \\
\leq &\frac{1}{n}(\|k_m(\cdot,x_j)k_{m'}(x_j,\cdot)\|_{\calH_m,\calH_{m'}} + \|k_m(\cdot,\tilde{x}_j)k_{m'}(\tilde{x}_j,\cdot))\|_{\calH_m,\calH_{m'}}) \notag \\
\leq &\frac{1}{n}(\|k_m(\cdot,x_j)\hnorm{m} \| k_{m'}(x_j,\cdot)\hnorm{m'} + \|k_m(\cdot,\tilde{x}_j)\hnorm{m}\|k_{m'}(\tilde{x}_j,\cdot))\hnorm{m'}) \notag \\
\leq &\frac{1}{n}(\sqrt{k_m(x_j,x_j)k_{m'}(x_j,x_j)} + \sqrt{k_m(\tilde{x}_j,\tilde{x}_j)k_{m'}(\tilde{x}_j,\tilde{x}_j)}) \notag \\
\leq &\frac{2}{n},
\label{eq:McDifferenceFirstTerm}
\end{align}
where we used $\|k_m(\cdot,x_j) \hnorm{m} = \sqrt{\langle k_m(\cdot,x_j),k_m(\cdot,x_j) \rangle_{\calH_m} } =\sqrt{k_m(x_j,x_j)}  $.
Bounding the norm of \eqref{eq:McDifferenceBasic} by 
\eqref{eq:McDifferenceFirstTerm},
we have 
\[
\|\hSigma_{m ,m'} - \Sigma_{m ,m'}\|_{\calH_m,\calH_{m'}}
-
\|\tSigma_{m ,m'} - \Sigma_{m ,m'}\|_{\calH_m,\calH_{m'}}
\leq \frac{2}{n}.
\]
By symmetry, changing $\hSigma$ and $\tSigma$ gives
\[
|\|\hSigma_{m ,m'} - \Sigma_{m ,m'}\|_{\calH_m,\calH_{m'}}
-
\|\tSigma_{m ,m'} - \Sigma_{m ,m'}\|_{\calH_m,\calH_{m'}}|
\leq \frac{2}{n}.
\]
Therefore 
by McDiarmid's inequality we obtain 
\begin{align*}
&P\left(\|\hSigma_{m ,m'} - \Sigma_{m ,m'}\|_{\calH_m,\calH_{m'}} - \EE[\| \hSigma_{m ,m'} - \Sigma_{m ,m'}\|_{\calH_m,\calH_{m'}}] \geq \varepsilon\right) \\
\leq &\exp\left(-\frac{2\varepsilon^2}{n(2/n)^2}\right) = \exp\left(-\frac{\varepsilon^2n}{2}\right).
\end{align*}
This gives the first assertion \Eqref{eq:SigmaConvFirstAss}.

To show the second assertion (\Eqref{eq:SigmaConvSecondAss}), first we note that 
\begin{align}
\EE[\| \hSigma_{m ,m'} - \Sigma_{m ,m'}\|_{\calH_m,\calH_{m'}} ] 
&\leq 
\sqrt{\EE[\| \hSigma_{m ,m'} - \Sigma_{m ,m'}\|_{\calH_m,\calH_{m'}}^2 ]} \notag \\
&= 
\sqrt{\EE[\| (\hSigma_{m ,m'} - \Sigma_{m ,m'})  (\hSigma_{m',m} - \Sigma_{m',m}) \|_{\calH_m,\calH_{m}} ]} \notag \\
&\leq 
\sqrt{\EE[\| (\hSigma_{m ,m'} - \Sigma_{m ,m'})  (\hSigma_{m',m} - \Sigma_{m',m}) \|_{\mathrm{tr}} ]}, 
\label{eq:EhSigmaSigmaBound}
\end{align}
where $\|\cdot\|_{\mathrm{tr}}$ is the trace norm and the last inequality. 
As in Lemma 1 of \cite{ALT:Gretton+etal:2005}, we see that 
\begin{align*}
&\| (\hSigma_{m ,m'} - \Sigma_{m ,m'})  (\hSigma_{m',m} - \Sigma_{m',m}) \|_{\mathrm{tr}} \\
=& \frac{1}{n^2}\sum_{i,j=1}^n \|k_m(\cdot,x_i) k_{m'}(x_i,x_j)k_m(x_j,\cdot)\|_{\mathrm{tr}} \\
&- \frac{2}{n} \sum_{i=1}^n \EE_X[ \|k_m(\cdot,x_i) k_{m'}(x_i,X)k_m(X,\cdot)\|_{\mathrm{tr}} ] + \EE_{X,X'}[ \|k_m(\cdot,X) k_{m'}(X,X')k_m(X',\cdot)\|_{\mathrm{tr}} ] \\
=&\frac{1}{n^2}\sum_{i,j=1}^n k_m(x_j,x_i) k_{m'}(x_i,x_j) - \frac{2}{n}\sum_{i=1}^n \EE_X[k_m(X,x_i) k_{m'}(x_i,X)] +  \EE_{X,X'}[ k_m(X',X) k_{m'}(X,X')],
\end{align*}
where $X$ and $X'$ are independent random variable distributed from $\Pi$.
Thus 
\begin{align*}
&\EE[\| (\hSigma_{m ,m'} - \Sigma_{m ,m'})  (\hSigma_{m',m} - \Sigma_{m',m}) \|_{\mathrm{tr}} ] \\
=&
\frac{n}{n^2} \EE_X[k_m(X,X) k_{m'}(X,X)] + \frac{n(n-1)}{n^2} \EE_{X,X'}[k_m(X',X) k_{m'}(X,X')]  \\
& - 2  \EE_{X,X'}[k_m(X',X) k_{m'}(X,X')] + \EE_{X,X'}[k_m(X',X) k_{m'}(X,X')] \\
=&\frac{1}{n} \EE_X[k_m(X,X) k_{m'}(X,X)] - \frac{1}{n} \EE_{X,X'}[k_m(X',X) k_{m'}(X,X')] \leq \frac{1}{n}.
\end{align*}
This and \Eqref{eq:EhSigmaSigmaBound} with the first assertion (\Eqref{eq:SigmaConvFirstAss}) gives the second assertion. 
\end{proof}


\begin{Lemma}
\label{lemm:noiseConvergeHspace}
If $\EE[\epsilon^2|X] \leq \sigma^2$ almost surely 
and $\sup_{X} k_m(X,X) \leq 1$, then 
we have 
\begin{align}
\| \hSigma_{m,\epsilon} \hnorm{m} = O_p(\sigma/\sqrt{n}).
\end{align}
\end{Lemma}
\begin{proof}
By definition, we have 
\begin{align*}
\EE[\| \hSigma_{m, \epsilon} \hnorm{m}]
& 
\leq \sqrt{\EE[\| \hSigma_{m,\epsilon} \hnorm{m}^2]} \\
& 
= \sqrt{\EE\left[ \frac{1}{n^2} \sum_{i,j=1}^n k_m(x_i,x_j) \epsilon_i \epsilon_j \right]} \\
& 
\leq \sqrt{ \frac{\sigma^2}{n}}.
\end{align*}
Applying Markov's inequality we obtain the assertion.
\end{proof}

\begin{proposition}[Bernstein's inequality in Hilbert spaces]
\label{prop:HilbertBernstein}
Let $(\Omega,\mathcal{A},P)$ be a probability space, $\calH$ be a separable Hilbert space, $\mathcal{B}>0$, and $\sigma>0$.
Furthermore, let $\xi_1,\dots,\xi_n:\Omega \to \calH$ be independent random variables satisfying $\EE[\xi_i]=0$, $\|\xi\|_{\calH}\leq B$, and 
$\EE[\|\xi_i\|_{\calH}^2]\leq \sigma^2$ for all $i=1,\dots,n$. 
Then we have 
\begin{align*}
P\left(\left\|\frac{1}{n}\sum_{i=1}^n \xi_i \right\|_{\calH} \geq \sqrt{\frac{2\sigma^2\tau}{n}} + \sqrt{\frac{\sigma^2}{n}} + \frac{2B\tau}{3n} \right)
\leq e^{-\tau}, ~~~~~(\tau >0).
\end{align*}
\end{proposition}
\begin{proof}
See Theorem 6.14 of \cite{Book:Steinwart:2008}.
\end{proof}

\bibliography{main,dalmkl}

\begin{thebibliography}{38}
\providecommand{\natexlab}[1]{#1}
\providecommand{\url}[1]{\texttt{#1}}
\expandafter\ifx\csname urlstyle\endcsname\relax
  \providecommand{\doi}[1]{doi: #1}\else
  \providecommand{\doi}{doi: \begingroup \urlstyle{rm}\Url}\fi

\bibitem[Bach et~al.(2004)Bach, Lanckriet, and Jordan]{ICML:Bach+etal:2004}
F.~Bach, G.~Lanckriet, and M.~Jordan.
\newblock Multiple kernel learning, conic duality, and the {SMO} algorithm.
\newblock In \emph{the 21st International Conference on Machine Learning},
  pages 41--48, 2004.

\bibitem[Bach(2008)]{JMLR:BachConsistency:2008}
F.~R. Bach.
\newblock Consistency of the group lasso and multiple kernel learning.
\newblock \emph{Journal of Machine Learning Research}, 9:\penalty0 1179--1225,
  2008.

\bibitem[Baker(1973)]{TAMS:Baker:1973}
C.~R. Baker.
\newblock Joint measures and cross-covariance operators.
\newblock \emph{Transactions of the American Mathematical Society},
  186:\penalty0 273--289, 1973.

\bibitem[Bartlett et~al.(2005)Bartlett, Bousquet, and
  Mendelson]{LocalRademacher}
P.~Bartlett, O.~Bousquet, and S.~Mendelson.
\newblock Local {R}ademacher complexities.
\newblock \emph{The Annals of Statistics}, 33:\penalty0 1487--1537, 2005.

\bibitem[Bickel et~al.(2009)Bickel, Ritov, and Tsybakov]{AS:Bickel+etal:2009}
P.~J. Bickel, Y.~Ritov, and A.~B. Tsybakov.
\newblock Simultaneous analysis of {L}asso and {D}antzig selector.
\newblock \emph{The Annals of Statistics}, 37\penalty0 (4):\penalty0
  1705--1732, 2009.

\bibitem[Bousquet(2002)]{BousquetBenett}
O.~Bousquet.
\newblock A {B}ennett concentration inequality and its application to suprema
  of empirical process.
\newblock \emph{C. R. Acad. Sci. Paris Ser. I Math.}, 334:\penalty0 495--500,
  2002.

\bibitem[Caponnetto and {de Vito}(2007)]{FCM:Caponetto+Vito:2007}
A.~Caponnetto and E.~{de Vito}.
\newblock Optimal rates for regularized least-squares algorithm.
\newblock \emph{Foundations of Computational Mathematics}, 7\penalty0
  (3):\penalty0 331--368, 2007.

\bibitem[Cortes(2009)]{ICMLtalk:Cortes:2009}
C.~Cortes.
\newblock Can learning kernels help performance?, 2009.
\newblock Invited talk at International Conference on Machine Learning (ICML
  2009). Montr{\'e}al, Canada, 2009.

\bibitem[Cortes et~al.(2009)Cortes, Mohri, and
  Rostamizadeh]{UAI:Cortes+etal:2009}
C.~Cortes, M.~Mohri, and A.~Rostamizadeh.
\newblock {$L_2$} regularization for learning kernels.
\newblock In \emph{the 25th Conference on Uncertainty in Artificial
  Intelligence (UAI 2009)}, 2009.
\newblock Montr{\'e}al, Canada.

\bibitem[Devroye et~al.(1996)Devroye, Gy{\"o}rfi, and
  Lugosi]{Book:ProbTheoryDevroye:1996}
L.~Devroye, L.~Gy{\"o}rfi, and G.~Lugosi.
\newblock \emph{A Probabilistic Theory of Pattern Recognition}.
\newblock Springer, 1996.

\bibitem[Gretton et~al.(2005)Gretton, Bousquet, Smola, and
  Sch{\"o}lkopf]{ALT:Gretton+etal:2005}
A.~Gretton, O.~Bousquet, A.~Smola, and B.~Sch{\"o}lkopf.
\newblock Measuring statistical dependence with {H}ilbert-{S}chmidt norms.
\newblock In S.~Jain, H.~U. Simon, and E.~Tomita, editors, \emph{Algorithmic
  Learning Theory}, Lecture Notes in Artificial Intelligence, pages 63--77,
  Berlin, 2005. Springer-Verlag.

\bibitem[Jia and Yu(2010)]{StatSinica:Jia+Yu:2010}
J.~Jia and B.~Yu.
\newblock On model selection consistency of the elastic net when p $\gg$ n.
\newblock \emph{Statistica Sinica}, 20\penalty0 (2):\penalty0 to appear, 2010.

\bibitem[Kloft et~al.(2009)Kloft, Brefeld, Sonnenburg, Laskov, M{\"u}ller, and
  Zien]{NIPS:Marius+etal:2009}
M.~Kloft, U.~Brefeld, S.~Sonnenburg, P.~Laskov, K.-R. M{\"u}ller, and A.~Zien.
\newblock Efficient and accurate $\ell_p$-norm multiple kernel learning.
\newblock In \emph{Advances in Neural Information Processing Systems 22}, pages
  997--1005, Cambridge, MA, 2009. MIT Press.

\bibitem[Koltchinskii(2006)]{Koltchinskii}
V.~Koltchinskii.
\newblock Local {R}ademacher complexities and oracle inequalities in risk
  minimization.
\newblock \emph{The Annals of Statistics}, 34:\penalty0 2593--2656, 2006.

\bibitem[Koltchinskii and Yuan(2008)]{COLT:Koltchinskii:2008}
V.~Koltchinskii and M.~Yuan.
\newblock Sparse recovery in large ensembles of kernel machines.
\newblock In \emph{Proceedings of the Annual Conference on Learning Theory},
  pages 229--238, 2008.

\bibitem[Lanckriet et~al.(2004)Lanckriet, Cristianini, Ghaoui, Bartlett, and
  Jordan]{JMLR:Lanckriet+etal:2004}
G.~Lanckriet, N.~Cristianini, L.~E. Ghaoui, P.~Bartlett, and M.~Jordan.
\newblock Learning the kernel matrix with semi-definite programming.
\newblock \emph{Journal of Machine Learning Research}, 5:\penalty0 27--72,
  2004.

\bibitem[Ledoux and Talagrand(1991)]{Book:Ledoux+Talagrand:1991}
M.~Ledoux and M.~Talagrand.
\newblock \emph{Probability in Banach Spaces. Isoperimetry and Processes}.
\newblock Springer, New York, 1991.
\newblock MR1102015.

\bibitem[Lin and Zhang(2006)]{AS:Lin+Zhang:2005:COSSO}
Y.~Lin and H.~H. Zhang.
\newblock Component selecion and smoothing in multivariate nonparametric
  regression.
\newblock \emph{The Annals of Statistics,}, 34\penalty0 (5):\penalty0
  2272--2297, 2006.

\bibitem[Meier et~al.(2009)Meier, {van de Geer}, and
  B{\"u}hlmann]{AS:Meier+Geer+Buhlmann:2009}
L.~Meier, S.~{van de Geer}, and P.~B{\"u}hlmann.
\newblock High-dimensional additive modeling.
\newblock \emph{The Annals of Statistics}, 37\penalty0 (6B):\penalty0
  3779--3821, 2009.

\bibitem[Mendelson(2002)]{IEEEIT:Mendelson:2002}
S.~Mendelson.
\newblock Improving the sample complexity using global data.
\newblock \emph{IEEE Transactions on Information Theory}, 48:\penalty0
  1977--1991, 2002.

\bibitem[Micchelli and Pontil(2005)]{JMLR:MicchelliPontil:2005}
C.~A. Micchelli and M.~Pontil.
\newblock Learning the kernel function via regularization.
\newblock \emph{Journal of Machine Learning Research}, 6:\penalty0 1099--1125,
  2005.

\bibitem[Rakotomamonjy et~al.(2008)Rakotomamonjy, Bach, Canu, and
  Y.]{JMLR:Rakotomamonjy+etal:2008}
A.~Rakotomamonjy, F.~Bach, S.~Canu, and G.~Y.
\newblock Simple{MKL}.
\newblock \emph{Journal of Machine Learning Research}, 9:\penalty0 2491--2521,
  2008.

\bibitem[Sonnenburg et~al.(2006)Sonnenburg, R{\"a}tsch, Sch{\"a}fer, and
  Sch{\"o}lkopf]{JMLR:Sonnenburg+etal:2006}
S.~Sonnenburg, G.~R{\"a}tsch, C.~Sch{\"a}fer, and B.~Sch{\"o}lkopf.
\newblock Large scale multiple kernel learning.
\newblock \emph{Journal of Machine Learning Research}, 7:\penalty0 1531--1565,
  2006.

\bibitem[Steinwart(2008)]{Book:Steinwart:2008}
I.~Steinwart.
\newblock \emph{Support Vector Machines}.
\newblock Springer, 2008.

\bibitem[Steinwart et~al.(2009)Steinwart, Hush, and
  Scovel]{COLT:Steinwart+etal:2009}
I.~Steinwart, D.~Hush, and C.~Scovel.
\newblock Optimal rates for regularized least squares regression.
\newblock In \emph{Proceedings of the Annual Conference on Learning Theory},
  pages 79--93, 2009.

\bibitem[Stone(1974)]{JRSS:Stone:1974}
M.~Stone.
\newblock Cross-validatory choice and assessment of statistical predictions.
\newblock \emph{Journal of the Royal Statistical Society, Series B},
  36:\penalty0 111--147, 1974.

\bibitem[Suzuki and Tomioka(2009)]{arXiv:Suzuki:2009}
T.~Suzuki and R.~Tomioka.
\newblock Spicymkl, 2009.
\newblock arXiv:0909.5026.

\bibitem[Talagrand(1996{\natexlab{a}})]{Talagrand1}
M.~Talagrand.
\newblock A new look at independence.
\newblock \emph{The Annals of Statistics}, 24:\penalty0 1--34,
  1996{\natexlab{a}}.

\bibitem[Talagrand(1996{\natexlab{b}})]{Talagrand2}
M.~Talagrand.
\newblock New concentration inequalities in product spaces.
\newblock \emph{Inventiones Mathematicae}, 126:\penalty0 505--563,
  1996{\natexlab{b}}.

\bibitem[Tomioka and Suzuki(2010)]{arXiv:SparsityTradeoff:2010}
R.~Tomioka and T.~Suzuki.
\newblock Sparsity-accuracy trade-off in {MKL}, 2010.
\newblock arXiv:1001.2615.

\bibitem[van~de Geer(2000)]{Book:VanDeGeer:EmpiricalProcess}
S.~van~de Geer.
\newblock \emph{Empirical Processes in {M}-Estimation}.
\newblock Cambridge University Press, 2000.

\bibitem[van~der Vaart and Wellner(1996)]{Book:VanDerVaart:WeakConvergence}
A.~W. van~der Vaart and J.~A. Wellner.
\newblock \emph{Weak Convergence and Empirical Processes: With Applications to
  Statistics}.
\newblock Springer, New York, 1996.

\bibitem[Vapnik(1998)]{book:Vapnik:1998}
V.~N. Vapnik.
\newblock \emph{Statistical Learning Theory}.
\newblock Wiley, New York, 1998.

\bibitem[Yuan and Lin(2007)]{JRSS:YuanLin:2007}
M.~Yuan and Y.~Lin.
\newblock On the nonnegative garrote estimator.
\newblock \emph{Journal of the Royal Statistical Society B}, 69\penalty0
  (2):\penalty0 143--161, 2007.

\bibitem[Zhang(2009)]{AS:TZhang:2009}
T.~Zhang.
\newblock Some sharp performance bounds for least squares regression with $l_1$
  regularization.
\newblock \emph{The Annals of Statistics}, 37\penalty0 (5):\penalty0
  2109--2144, 2009.

\bibitem[Zhao and Yu(2006)]{JMLR:ZhaoYu:2006}
P.~Zhao and B.~Yu.
\newblock On model selection consistency of lasso.
\newblock \emph{Journal of Machine Learning Research}, 7:\penalty0 2541--2563,
  2006.

\bibitem[Zou and Hastie(2005)]{JRSS:Zou+Hastie:2005}
H.~Zou and T.~Hastie.
\newblock Regularization and variable selection via the elastic net.
\newblock \emph{Journal of the Royal Statistical: Series B}, 67\penalty0
  (2):\penalty0 301--320, 2005.

\bibitem[Zou and Zhang(2009)]{AS:Zou+Zhang:2009}
H.~Zou and H.~H. Zhang.
\newblock On the adaptive elastic-net with a diverging number of parameters.
\newblock \emph{The Annals of Statistics}, 37\penalty0 (4):\penalty0
  1733--1751, 2009.

\end{thebibliography}

\end{document}